# Online Robot Motion Planning Methodology Guided by Group Social Proxemics Feature

Xuan Mu, Xiaorui Liu, Shuai Guo, Wenzheng Chi, Wei Wang and Shuzhi Sam Ge, *Fellow, IEEE*

***Abstract*—Nowadays robot is supposed to demonstrate human-like perception, reasoning and behavior pattern in social or service application. However, most of the existing motion planning methods are incompatible with above requirement. A potential reason is that the existing navigation algorithms usually intend to treat people as another kind of obstacle, and hardly take the social principle or awareness into consideration. In this paper, we attempt to model the proxemics of group and blend it into the scenario perception and navigation of robot. For this purpose, a group clustering method considering both social relevance and spatial confidence is introduced. It can enable robot to identify individuals and divide them into groups. Next, we propose defining the individual proxemics within magnetic dipole model, and further established the group proxemics and scenario map through vector-field superposition. On the basis of the group clustering and proxemics modeling, we present the method to obtain the optimal observation positions (OOPs) of group. Once the OOPs grid and scenario map are established, a heuristic path is employed to generate path that guide robot cruising among the groups for interactive purpose. A series of experiments are conducted to validate the proposed methodology on the practical robot, the results have demonstrated that our methodology has achieved promising performance on group recognition accuracy and path-generation efficiency. This concludes that the group awareness evolved as an important module to make robot socially behave in the practical scenario.**

***Note to practitioners* — For robot cruising in the dynamic and interactive scenarios, it is supposed to actively perceive people and initiate the interaction with them. So, for the navigation module of robot, it not only needs to consider obstacle avoidance, but also contribute to facilitating the interaction between robot and people. In this paper, we hope making robot has human-like social awareness in two aspects. Firstly, robot is capable to identify whether people are part of a group; secondly, robot can search and determine the position to observe or interact with specific group. For this purpose, we establish the clustering method to recognize the group in the public scenario, and present representing the proxemics of individual and group within vectorized field model. Since the proxemics space of group (local map) and scenario (global map) are both unique plane field, its mixture feature can naturally help robot determine the cruising path. The experimental results have offered the possibilities to realize the interaction-oriented navigation on the robot, and achieve satisfactory performance in dynamic scenario.**

Manuscript received 9 January 2024. This work was supported in part by the Shandong Provincial Natural Science Foundation under Grant ZR2023QF002, in part by the Qingdao Natural Science Foundation under Grant 23-2-1-126-ZYYD-JCH, in part by the National Key Research and Development Program of China under Grant 2020YFB1313600 (Corresponding author: Xiaorui Liu)

Xuan Mu is with the School of Automation, Qingdao University, Qingdao 266000, China (e-mail: muxuanqdu@163.com).

Xiaorui Liu is with the School of Automation, Qingdao University, Qingdao 266000, China (e-mail: liuxiaorui@qdu.edu.cn).

Shuai Guo is with the School of Automation, Qingdao University, Qingdao University, Qingdao 266000, China (e-mail: guoshuai7212@163.com).

Wenzheng Chi is with Robotics and Microsystems Center, School of Mechanical and Electric Engineering, Soochow University, Suzhou 215021, China (e-mail: wzchi@suda.edu.cn).

Wei Wang is with the School of Mechanical Engineering & Automation and the Robotics Institute, Beihang University, Beijing 100000, China (e-mail: wangweilab@buaa.edu.cn).

Shuzhi Sam Ge is with the Department of Electrical and Computer Engineering, National University of Singapore, Singapore 117576, on leave from the School of Automation, Qingdao University, Qingdao 266000, China (email: elegesz@nus.edu.sg).

## I. INTRODUCTION

Nowadays, various robots have been employed in a wide range of application [1,2]. Since putting robots into human environment necessitates addressing social interaction, there is practical need to investigate comprehensive, quantitative understanding on the mechanism of human social behavior [3, 4]. In [5], Broda M D *et al*. have shown that individual face fixation biases are at least partly driven by domain-general active vision. Camara, F *et al*. analyze the non-circular boundary of pedestrian, and proposed a proxemics model to represent its kinematics features [6]. Based on above discoveries about human social behavior patterns, several robot social-oriented frameworks have been presented recently. In [7], K Cai *et al*. present a human-aware path planner which considers collision risk and dynamic group as constrains, and employ an improved virtual doppler method to prevent robot from falling into the deadlock area. Yuan Y *et al*. attempt to fuse the pedestrian density and obstacles in feature level, and propose allowing robotic systems to efficiently generate heuristic paths based on prior knowledge of environment mixture [8]. Although these studies have achieved promising advances to make robot behave like human, it is still a major challenge to blend social principles into the robot control [9].

As for the social principles, one of the core requirements is the representation of the human group [10]. Up to now, the identification of the human group has reached the basically satisfactory performance. Such as Lu X Y *et al*. propose a siamese transformer for group re-identification, integrating multiscale recognition cores and joint learning module to realize the robust representations [11]. Zhou C *et al*. develop a social group model that describes a front−back asymmetric social interaction field. It could predict participants' perceptual judgments of social grouping in static and dynamic scenes [12]. On the basis of the group identification technologies, another problem is how to model the topological space of social groups (*i.e.* proxemics [13]). Comparing with the identification tasks, the studies of proxemics pay more attention to build the interaction between individuals or groups, and how people within groups influence each other [14]. For robot being applied in dynamic and long-term social scenario, it needs to not only recognize the human group, but also arrange the own behavior control (such as path planning, gaze shift, pose adjustment etc.) based on their proxemics [15-16]. In existing studies, people usually intend to make robot maintain comfortable distances from people to avoid physical and psychological invasion. Carlos M S *et al*. consider that the context surrounding robots and persons affects the expected behavior, defines an adaptive proxemics area around a person that adapts to the real situation [17]. Bozorgi H *et al*. combine human content presented by Hall's Proxemics model with the robot kinematics. They present a kind of parameter, social dynamic confidence, to determine the robot path based on features of spatial norms, dynamic metrics and consistency developed in the detection stage [18]. Besides, there are some studies on the non-geometric proxemics representation [19], such as Camara F *et al*. proposing one continuous pedestrian space functions to describe pedestrian invasion case [20]; Sousa R M *et al*. employing asymmetric Gaussian function (for individuals and groups) to define the zones that individual or group occupies [21].

Based on the representation of the social space, it need complete the robot motion planning among group topology or superposition field [22, 23]. In [24, 25], Yao W *et al*. discuss the robot path-planning and kinematics control problem in the arbitrary vector field, and propose a method to obtain the domain of attraction of the desired path. Although these works have basically addressed the existence of singular points (where the vector field vanishes) and given the solution to deal with the convergence obstruction of desired path, but the computation based on geometric solving is still complex [26]. Moreover, if considering the differential constrains and underactuated control of robot, this problem will become further complicated. Under these circumstances, the sampling-based (or heuristic) algorithm is more feasible to obtain near-optimal path through efficient configuration space modeling and global state searching, which make it become popular for a very general class of problems [27]. Zucker *et al*. proposed the multipartite rapidly exploring random tree algorithm (RRT) that supports the motion planning in unknown or dynamic environments [28]. Palmieri *et al*. present a mobile robot motion planning approach under kinodynamic constraints that exploits learned perception priors in the form of continuous Gaussian mixture fields [29]. Summers et al. build a distributionally robust incremental sampling-based method for motion planning under uncertainty, which can explicitly incorporate localization error, stochastic process disturbances and unpredictable obstacle motion [30].

Motivated by above works, this paper primarily focuses on modeling the proxemics of individual and group, and investigates how to blend into the robot motion planning in dynamic social scenario. The main contribution of this paper contains following aspects:

1) Building the group awareness of robot, and developing a novel group clustering method that considers both social relevance (SR) and spatial confidence (SC). Within this method, robot can identify whether people are part of a group.
2) Based on the group clustering, propose defining the proxemics of group within magnetic diploe model. For the vectorized proxemics field of single group, a method to calculate its optimal observation positions (OOPs) is established.
3) Based obtained OOPs grid and proxemics map, propose a sampling-based and hierarchical path planning algorithm for the robot in the scenario composed of multiple groups and static obstacle. This algorithm contains the combination optimization solver to determine group observation sequence and sampling-based path planning module to generate segmented cruise path. Both simulation and physical experiments have been conducted based on service robot platform, proving its feasibility to blend the group awareness into robot interactive control.

The rest of this article is organized as follows. Section II introduces the methods regarding group interaction space and optimal observation positions calculation, and discusses the path-planning method to lead robot in dynamic proxemics field. In Section III, the simulation and result analysis for proposed methodology is demonstrated. In Section IV, it contains the physical experiment setup and discussion. Finally, Section V concludes this article.

## II. Methodology

For the robot operating in the open environment, it primarily needs to perceive the various social groups and build the representation of their proxemics space. The proposed framework in this article is shown as Fig. 1. It contains three functional modules: social group awareness module, proxemics field superposition module, and interactive navigation module. It is assumed that the motion state of the individuals $p_m^n(t)$, robot $p_r(t)$, and obstacles $(OBS)$ in given scenario can be obtained through the motion capture system. For each individual or robot, the motion state contains two parameters: position coordination $\{x(t), y(t)\}$ and heading direction $\theta(t)$. $\mathcal{G}_m^n(t)$ denotes the *m*th social group, it is a flexible matrix to contains the motion states of all group member $p_m^1(t) \dots . p_m^n(t)$. Since there is requirement to design a boundary for $obs_k$, we define the obstacle boundary $\mathcal{O}_\mathrm{k}$ which is the real border of $obs_k$, and $\mathcal{B}_k$ represents the reactive boundary for $\mathrm{obs_k}$. $\mathcal{I}_{\mathcal{O}_k}$ and $\mathcal{I}_{\mathcal{B}_k}$ are line integrals of $\mathcal{O}_k$ and $\mathcal{B}_k$respectively. In order to ensure obstacle avoidance during robot motion, an assumption needs to be presented (there holds $\mathcal{I}_{\mathrm{O_k}} \subseteq \mathcal{I}_{\mathcal{B}_\mathrm{k}}$). In social group awareness module, robot collects the states of the people, and determine whether people are part of a group via clustering calculation. Through identifying people into various groups ($\mathcal{G}_1 \sim \mathcal{G}_4$ in Fig. 1), the basic group awareness of robot can be established.

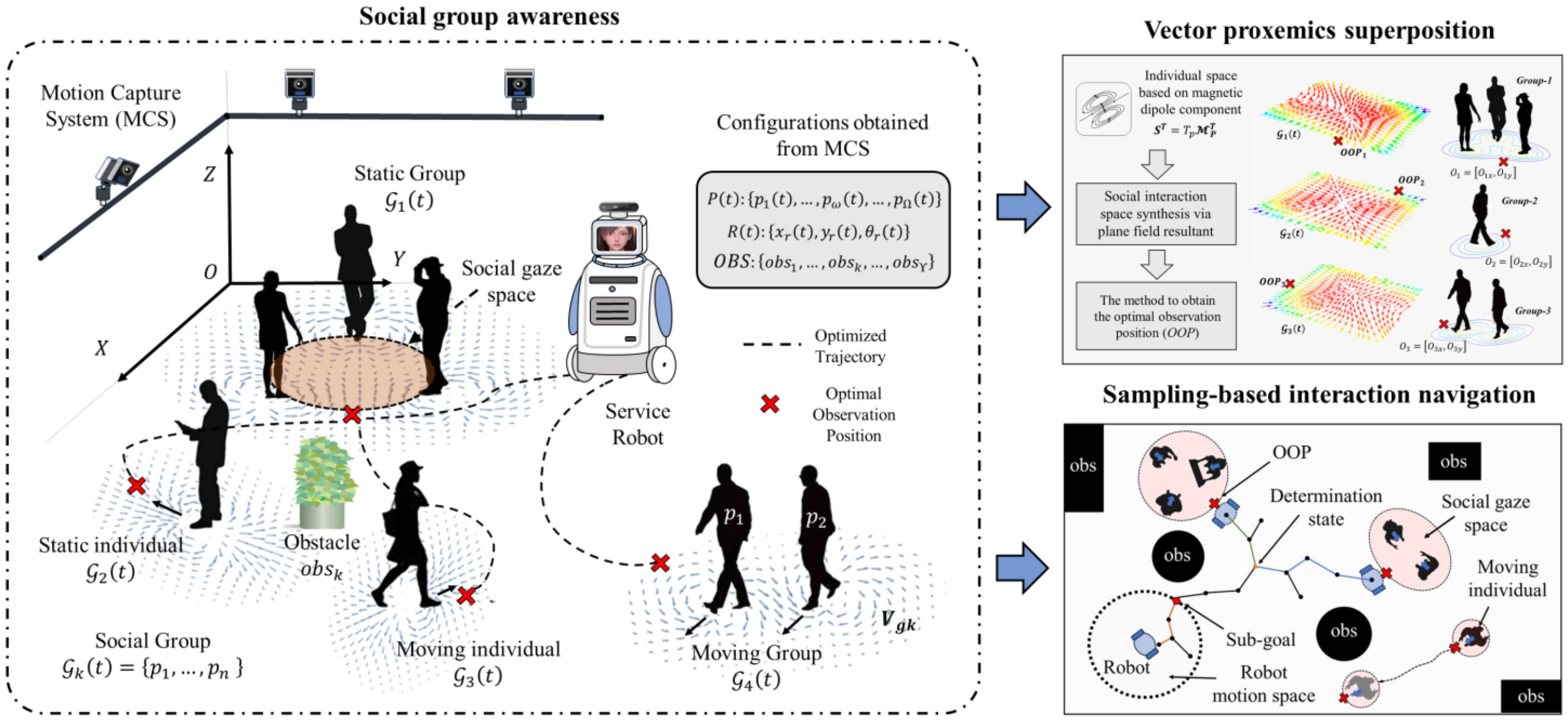

Fig. 1. System diagram of proposed robot control framework. Within the Groups Awareness Module, based on the data from MCS, the individuals are classified into distinct groups. Combined with the presented social space model, the Optimal Observation Position is computed by the constraints according to social norms. The Robot Behavior Driver Module is tasked with generating a comfortable path to approach groups while avoiding obstacle. And the details of our work are described below.

Based on the states of various groups, a social vector field $V_g$ is built (light cyan arrows in Fig. 1) to represent their proxemics space. Partially motivated by artificial potential field (APF) theory [31], [32], a vector field model based on magnetic dipole component is employed to describe the proxemics space of individual people. Through superposing the proxemics spaces of group members, the total proxemics spaces of group can be obtained. Through further superposing the proxemics spaces of groups, the global map with can be established. Within above methods, robot could transfer the group with any size (including individual) or topology into the unique two-dimension (2D) vector field, which can provide convenience for global scenario understanding. In previous studies, people usually intend to regard people or group as some kind of obstacle. As a result, the proxemics models are employed more to guide robot for avoidance. Although this enables robot to cruise in the crowed environment, it cannot facilitate the practical interaction between robot and people (*i.e.* obstacle). Therefore, in vector proxemics superposition module, we propose utilizing 2D vector field to represent group proxemics, and defining the concept of the optimal observation position (OOP) ($O_k$, red cross in Fig. 1). OOP represents the ideal position for robot to observe or interact with the given group. It is calculated based on flux density of $V_{gk}$ of $\mathcal{G}_k$. On the other hands, we quote the concept of social gaze space that indicates the group members have mutual eye contact and prohibit external disruption [33, 34]. On the basis of the global proxemics space and group feature representation, sampling-based navigation module formulates the path in robot motion space (black dotted circle in the lower grey box of Fig. 1). For robot operating in the scenarios containing various groups, this module firstly calculates the proxemics spaces, OOPs and social gaze space (light red circle in the lower grey box of Fig. 1). Then, the visiting sequence among these groups will be determined based on combinational optimization method. Through determining group observation sequence, the global cruising path can be divided into several quasi-optimal path along with the OOPs grid. When the robot approaches a specific group, a quasi-optimal path between robot and specific group is generated. Then, integrated with robot motion space, sub-goal is formulated and robot will arrive at sub-goal. The brown lines in Fig. 1 are portions of the previously generated quasi-optimal path. When sub-goal reaches the determination state, the robot would determine which group is more deserving to approach, green lines and blue lines indicate the paths to different groups. According to the obtained cruising path, robot will try to visit all group in the proxemics-OOPs system. When reaching an OOP, robot can observe this group, then decides whether initiate the interaction with its members. The detailed proposed framework will be introduced in three parts. The details of above work will be discussed in the following parts.

### *A. Group Clustering based on multiple social factors*

The clustering method is to classify the individuals into static or dynamic social groups. Based on the features of proxemics, there are four types of clusters in social scenarios: static individual, dynamic individual, static group, dynamic group ($\mathcal{G}_1 \sim \mathcal{G}_4$ in Fig. 1). In the perspective of social psychology, it is supposed that closer interpersonal distances, more direct interpersonal angles and more open avatar postures will led to a higher probability of a group being judged as interactive [4]. Therefore, we propose an interaction field clustering algorithm fusing social relevance and spatial correlation, which identify participants' perceptual judgments of social grouping. Its formulation is as follows.

1) **Social relevance (SR)**: partially motivated by [12], a concept of social relevance is to describe the relation between two persons, namely $S_R$. For any two people, their interaction relationship is shown as Fig. 2. $p_A$ and $p_B$ are interacting with each other. It assumes $\boldsymbol{\Delta}$ as vector from $O_A = [x_{pA}, y_{pA}]$ to $O_B = [x_{pB}, y_{pB}]$, where $O_A$ and $O_B$ represent the coordinates of $p_A$ and $p_B$ respectively in terrestrial coordinate system. Besides, there is an interaction factors $I_{fk}$ to describe the social influence of person $p_k$. It represents the social impact irrelevant as the heading direction. Based on above definition, the social relevance $S_{RAB}$ between $p_A$ and $p_B$ integrated with interaction factor $I_{fA}$ and $I_{fB}$ can be defined as Eq. (1).

$$S_{RAB} = \left\| \frac{(I_{fA} \cdot \boldsymbol{\Delta} + cI_{fA})(I_{fB} \cdot -\boldsymbol{\Delta} + cI_{fB})}{\|\boldsymbol{\Delta}\|^2} \boldsymbol{\Delta} \right\|^2 \quad (1)$$

where $c$ is constant coefficient, $\|\boldsymbol{\Delta}\|^2$ denotes the Euclidean distance of $\boldsymbol{\Delta}$. $I_{fA} = I_{fB} = I$ represents the social impact equal between $p_A$ and $p_B$. Then, the social relevance can be reorganized in the form of heading angle, as Eq. (2-3).

$$\psi(\cos\theta) = \begin{cases} \cos\theta & if \cos\theta > 0 \\ 0 & otherwise \end{cases} \quad (2)$$

$$S_{RAB} = I^2 \frac{(\psi(\cos\theta_A) + c)(\psi(\cos\theta_B) + c)}{\|\boldsymbol{\Delta}\|^2} \quad (3)$$

where $\theta_A$ is angle between the heading of $p_A$ and $\boldsymbol{\Delta}$, $\theta_B$ is angle between the heading of $p_B$ and $-\boldsymbol{\Delta}$. Within above definition, the interaction correlation between two people can be evaluated. Next, the calculation about spatial confidence and group clustering will be conducted.

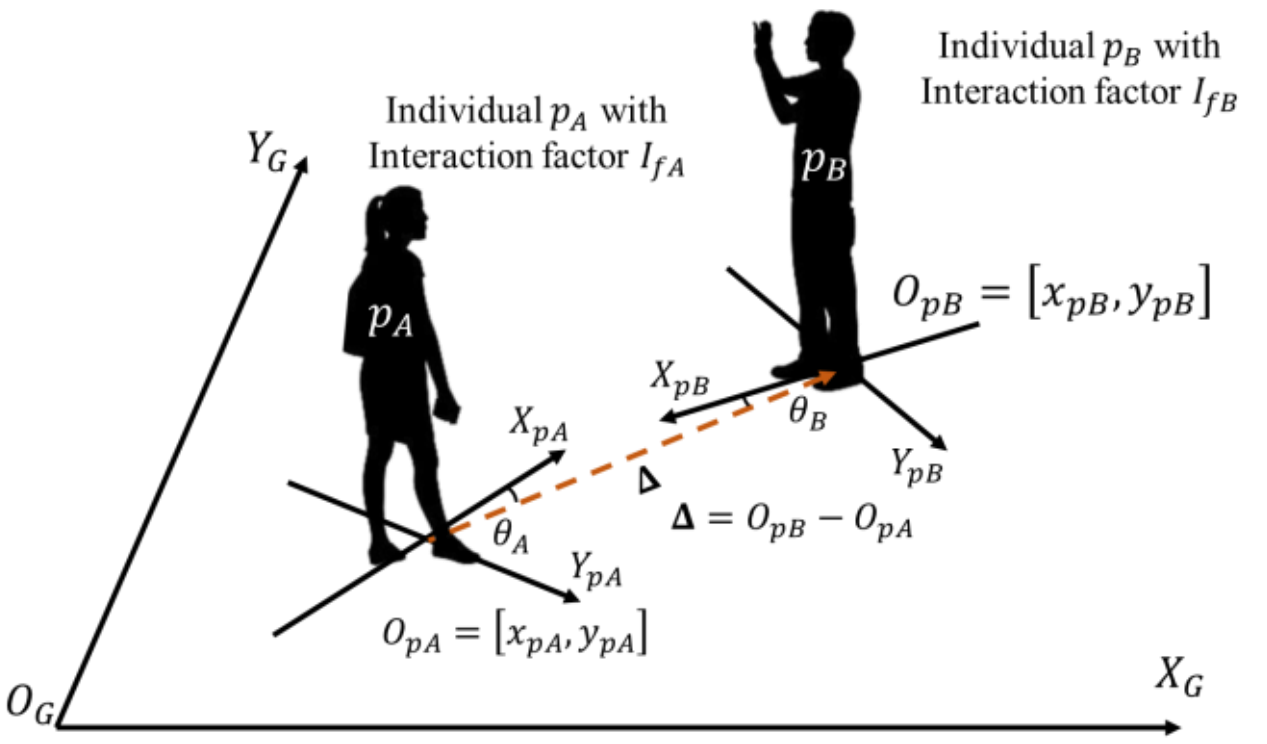

Fig. 2. The social relevance $S_R$ between $p_A$ and $p_B$.

**2) Spatial Confidence (SC)**: it is used to determine whether two people are part of a group. Here the Weibull distribution model is used to define SC ( $\eta$ ) as Eq. (4).

$$\eta = 1 - e^{-\left(\frac{S_R}{\lambda}\right)^b} \quad (4)$$

where $\lambda$ represent the threshold, we set $a = I^2/\lambda$. Then, the Eq. (4) can be transformed to Eq. (5).

$$\eta_{AB} = 1 - e^{-a\left(\frac{(\psi(\cos\theta_A)+c)(\psi(\cos\theta_B)+c)}{\|\boldsymbol{\Delta}\|^2}\right)^b} \quad (5)$$

where $a, b, c$ are constant parameter. It assumed that there are $n$ people present in the current social scenario. The set of $n$ people are described as $p^n(t) = \{p_1, \dots, p_n\}$, where $p_n(t) = [x_n, y_n, \theta_n]$. Then, the spatial confidence matrix $H$ can be obtained in the form of Eq. (6), it is the basis to complete grouping clustering.

$$H = \begin{bmatrix} \eta_{11} & \eta_{12} & \eta_{13} & \cdots & \eta_{1n} \\ \eta_{21} & \eta_{22} & \eta_{23} & \cdots & \eta_{2n} \\ \eta_{31} & \eta_{32} & \eta_{33} & \cdots & \eta_{3n} \\ \vdots & \vdots & \vdots & \vdots & \vdots \\ \eta_{n1} & \eta_{n2} & \eta_{n3} & \cdots & \eta_{nn} \end{bmatrix} \quad (6)$$

**3) Grouping Clustering**: within spatial confidence matrix, a clustering method based on graph connectivity is proposed to generate group set $\boldsymbol{\mathcal{G}} = \{\mathcal{G}_1, \dots, \mathcal{G}_m\}$. It is assumed that the social group topology can be described as $G = (\mathcal{V}, \mathcal{E}, \mathcal{A})$. It has a vertex set $\mathcal{V} = \{1, \dots, n\}$, an edge set $\mathcal{E} \subseteq \mathcal{V} \times \mathcal{V}$ and the weighted adjacency matrix $\mathcal{A} = [a_{ij}] \in \mathbb{R}^{\Omega\times\Omega}$ . $a_{ij}$ can be obtained by spatial confidence matrix H: $a_{ij} = \eta_{ij}$ if $i < j$ and $a_{ij} = 0$ otherwise. The detailed process of grouping clustering is shown in Fig. 3. The black circle with red number represents the vertex, the light bule line with double arrow is edge in graph, the black dotted circle represents the group clusters. $a_{ij}$ is weight of edge from vertex $i$ to vertex $j$. Then, it is assumed that $a_{thers}$ is the threshold of spatial confidence.

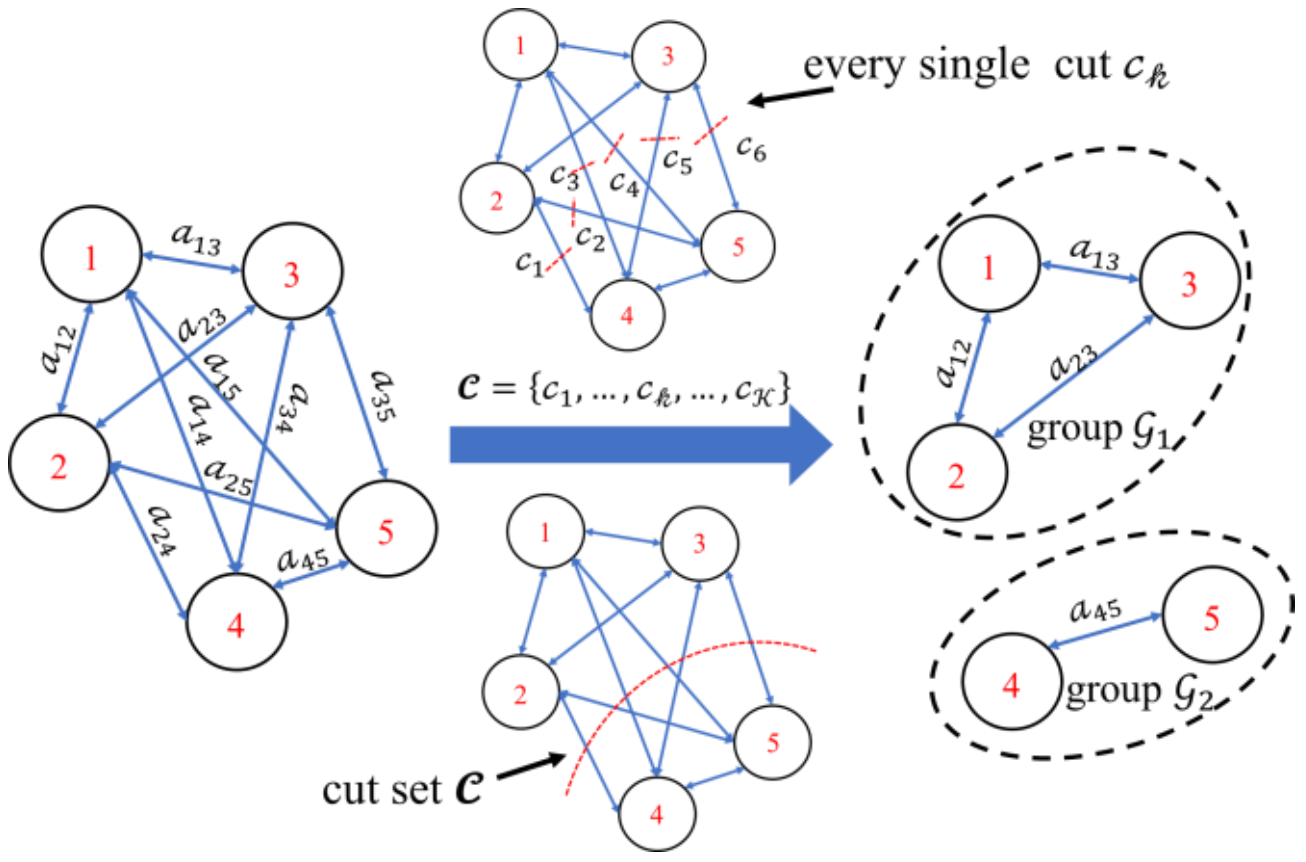

Fig. 3. The process of clustering based on graph connectivity of spatial confidence matrix $H$

To build the connectivity of graph $G$ based on spatial confidence, the cut set $\boldsymbol{\mathcal{C}}$ needs to be proposed to cut the edge whose weight is lower than $a_{thers}$. The red straight dotted lines represent the single cuts in graph. The cut set is defined as $\boldsymbol{\mathcal{C}} = \{c_1, \dots, c_k, \dots, c_{\mathcal{K}}\}$, where $c_k \leftarrow [i, j]$ if $a_{ij} < a_{thers}$. The edge with weight $a_{ij}$ (its subscript is stored in set $\boldsymbol{\mathcal{C}}$) will be removed from graph $G$. The red arc (dotted lines) represents the cut set in graph. The group set $\boldsymbol{\mathcal{G}}$ can be obtained by using the Depth First Search (DFS) algorithm combined with connectivity of graph $G$ Within above assumption, the group set $\boldsymbol{\mathcal{G}}$ can be utilized in observation position calculation and path planning.

*B. The magnetic dipole-social proxemics model and method to solve optimal observation positions*

To represent group proxemics and blend it into global map building, a mathematic model is needed to accurately describe the space around human group. Here we propose the social proxemics model based on magnetic dipole component and the concept of optimal observation position (OOP). Their definition and calculation method will be introduced as follows.

**1) Magnetic dipole proxemics model:** partially inspired by the formation mechanism of planetary magnetic field [35], we used the magnetic dipole model to represent the social space around the human. The two-dimensional magnetic dipole model $\boldsymbol{\mathcal{M}}$ is defined as Eq. (7-8). Since the magnetic field line always comes from positive pole to negative pole, it can naturally represent the front and back of human.

$$\mathcal{M}_x(x,y,x_i,y_i) = \alpha \frac{2(x-x_i)^2-(y-y_i)^2}{((x-x_i)^2+(y-y_i)^2)^{\frac{5}{2}}} \tag{7}$$

$$\mathcal{M}_y(x,y,x_i,y_i) = \alpha \frac{3(x-x_i)(y-y_i)}{((x-x_i)^2+(y-y_i)^2)^{\frac{5}{2}}} \tag{8}$$

In Eq. (7-8), α is constant coefficient, $\mathcal{M}_x$ and $\mathcal{M}_y$ is the magnetic flux density of $\boldsymbol{\mathcal{M}}$ in $X_G$ direction and $Y_G$ direction. $[x_i, y_i] \in \mathbb{R}^2$ represents the central coordinates of the vector field model $\boldsymbol{\mathcal{M}}$. It is assumed that $p = [x_p, y_p, \theta_p]$ is the basic human motion state, and $\boldsymbol{\mathcal{H}}$ represent the magnetic dipole model combined with $p$. Then, the formal magnetic dipole-social space model $\boldsymbol{S}$ is defined as Eq. (9-11).

$$\boldsymbol{\mathcal{H}} = \boldsymbol{\mathcal{M}}(x, y, x_p, y_p) \tag{9}$$

$$T_p = \begin{bmatrix} \cos\theta_p & -\sin\theta_p \\ \sin\theta_p & \cos\theta_p \end{bmatrix}^T \tag{10}$$

$$\boldsymbol{S}^T = T_p \boldsymbol{\mathcal{H}}^T \tag{11}$$

where $T$ denotes transpose of matrix, and $T_p$ represents the transformation matrix for $\boldsymbol{\mathcal{H}}$. The model is illustrated as Fig. 4.

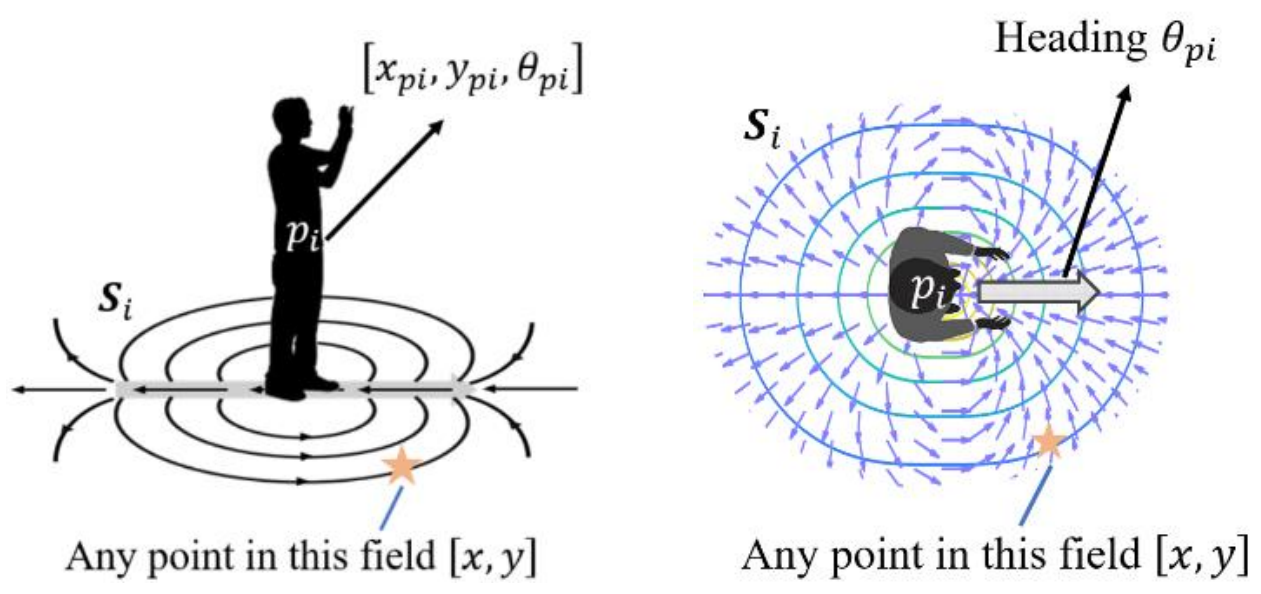

(a) (b)

Fig. 4. The schematic diagram of magnetic dipole proxemics space model. (a) $[x_{pi}, y_{pi}, \theta_{pi}]$ is configuration of human $p_i$ in social scene. $\boldsymbol{S_i}$ stands for this model. (b) shows details of space model, the contour and blue arrow represents the amplitude and direction of vector field respectively.

**2) The OOP calculation method**: the concept of OOP is proposed to designate a location to facilitate robot-group observation and interaction. Let $O_k$ stands for OOP for group $\mathcal{G}_k$, $k = 1, \dots, m$. The calculation method for determining the OOP contains proximity constraints and effects from the magnetic dipole proxemics model. To integrate the above two components, the algorithm is introduced: $\mathcal{G}_k = \{p_1, \dots, p_j, \dots, p_n\}$, $p_j = [x_{pj}, y_{pj}, \theta_{pj}]$ represents the group $\mathcal{G}_k$ contains $n$ individuals. Based on our proposed model, the social vector field $\boldsymbol{V_{gk}}$ for $\mathcal{G}_k$ can be expressed as Eq. (12).

$$\boldsymbol{V_{gk}} = \sum_{j=1}^{n} \boldsymbol{S}_j \tag{12}$$

where $\boldsymbol{S}_j$ denotes magnetic dipole proxemics space model of $p_j$ within group $\mathcal{G}_k$. To accurately determine the location of OOP, the definition of social damping is proposed. Let $\boldsymbol{\xi}_k \in \mathbb{R}^2$ represents social damping for $\mathcal{G}_k$. As show in Fig. 5, all light cyan arrows denote the vector field, the red cross stands for OOP and the yellow arrow represents social damping. The calculation integrated with $\boldsymbol{\xi}_k$, $O_k$ and $\boldsymbol{V_{gk}}$ is defined as Eq. (13-16).

$$\boldsymbol{V_{gk}}(O_k) + \boldsymbol{\xi}_k = \boldsymbol{0} \tag{13}$$

$$Amp(\boldsymbol{\xi}_k) = \beta \frac{e^{-n}}{n} \tag{14}$$

$$Arg(\boldsymbol{\xi}_k) = \frac{\sum_{j=1}^{n} \theta_j}{n} + \pi \tag{15}$$

$$\theta_j = \tan^{-1}\left(\frac{O_{k_y} - y_{pj}}{O_{k_x} - x_{pj}}\right) \tag{16}$$

where $n$ is the size of $\mathcal{G}_k$, and $\beta$ is coefficient of social damping. $\theta_j$ is the angel between positive $X_G$ direction and the line connecting $O_k$ with $[x_{pj}, y_{pj}]$. $Amp$ and $Arg$ are function to calculate the amplitude and direction of social damping respectively. Based on equation (13), the OOP can be obtained based on group proxemics fields. With the generation of OOP, the social gaze space is formulated. In Fig. 5 (a) (b) (c), the red dotted circle represents the range of social gaze space, the blue round represents group gaze center $\gamma_k$ and little blue round represents gaze point of $p_j$. Let $\gamma_{j,k}$ denotes gaze point for $p_j$. $\gamma_k$ and $\gamma_{j,k}$ can be expressed as:

$$\gamma_{j,k_x} = x_{pj} + r\cos\theta_{pj} \tag{17}$$

$$\gamma_{j,k_y} = y_{pj} + r\sin\theta_{pj} \tag{18}$$

$$\gamma_k = \frac{\sum_{j=1}^{n} \gamma_{j,k}}{n} \tag{19}$$

where $r$ is constant coefficient. Then, the radius of social gaze space $\mathcal{R}_k = \|\gamma_k - O_k\|^2$. The concept of social gaze space is conducive to conduct path-planning in next part for robot in social scenes.

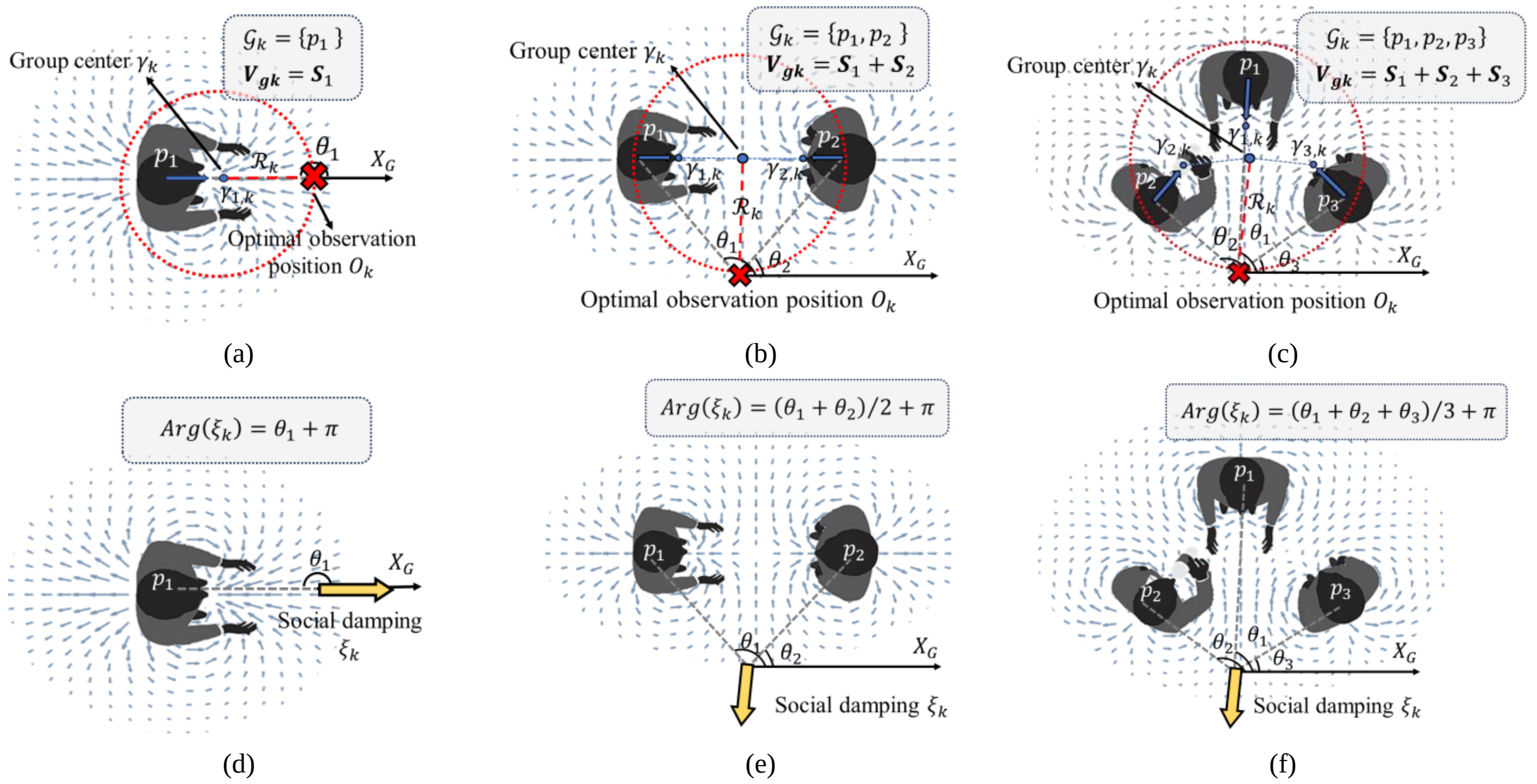

Fig. 5. The schematic diagram of optimal observation position, social gaze space and social damping.

*C. Sampling-based path planning under global proxemics field*

Incorporated with the social vector field $\boldsymbol{V_{gk}}$ and classical RRT* algorithm [26], a path planning method is proposed to satisfy the requests from environments and proxemics principles.

**1) Basic definition and configuration**: let $\mathcal{S} \in \mathbb{R}^2$ stand for the state space, $\mathcal{S}_{obs}$ is the space of obstacle and $\mathcal{S}_{free}$ represents free space, where $\mathcal{S}_{obs} = \mathcal{I}_{\mathcal{B}_k}$. Provided the initial state $\mathcal{s}_{start} \in \mathcal{S}_{free}$ and goal state $\mathcal{s}_{goal} \in \mathcal{S}_{goal}$, $\mathcal{S}_{goal}$ denotes the goal space and $\mathcal{S}_{goal} \in \mathcal{S}_{free}$. The essence in motion planning motion planning for robot is to find a feasible motion planning for robot is to find a feasible path $\mathcal{V} = \mathcal{s}(\tau) \in \mathcal{S}_{free}$, where $\tau \in [0, T], \mathcal{s}(0) = \mathcal{s}_{start}$ *and* $\mathcal{s}(T) \in \mathcal{S}_{goal}$. In RRT* algorithm, a tree $\Gamma$ is defined to realize connection between $\mathcal{s}_{start}$ and $\mathcal{s}_{goal}$ where $\mathcal{s}_{start}$ is the root of the tree $\Gamma$ (the terminal state belongs to $\mathcal{S}_{goal}$). Here, $\mathcal{S}_{SGS}$ represents the social gaze space of groups in scene, and the relation with $\mathcal{S}_{free}$ and $\mathcal{S}_{obs}$ is expressed as $\mathcal{S}_{free} = \mathcal{S} \backslash (\mathcal{S}_{obs} + \mathcal{S}_{SGS})$. The random sampling method in RRT* carries out in $\mathcal{S}_{free}$.

**2) Vector field of magnetic dipole-RRT* (VMD-RRT*)**: the main idea of VMD-RRT* is to employ RRT* path planning method to identify a reasonable path guided by superpositional social vector field $\boldsymbol{V_{gC}} = \sum_{k=1}^{N} \boldsymbol{V_{gk}}$. Algorithm 1 outlines the workflow of VMD-RRT*. The inputs of Algorithm 1 is $\boldsymbol{u_{vmd}} = [\mathcal{s}_{start}, \mathcal{s}_{goal}, \mathcal{S}_{goal}, M_S, \varepsilon, \boldsymbol{\mathcal{F}_{md}}, r]$, where $\mathcal{s}_{start}$ is initial state, $\mathcal{s}_{goal}$ denotes goal state, $\mathcal{S}_{goal}$ is goal space, $M_S$ indicates the map, $\varepsilon$ is the coefficient in Steer function, $\boldsymbol{\mathcal{F}_{md}}$ is the effects from vector field of Magnetic Dipole, $r$ denotes the searching radius. The output of VMD-RRT* is a tree $\Gamma$. Specifically, $\Gamma$ is initialized with $\mathcal{s}_{start}$ as the root (Line 1 of Algorithm 1). A random sampling method is utilized to generate candidate state $\mathcal{s}_{rand}$ in map $M_S$ during exploring process. Then, the closest state $\mathcal{s}_{nearest}$ to $\mathcal{s}_{rand}$ in current path $\mathcal{V}$ is chosen. The new state $\mathcal{s}_{new}$ is generated by function Steer [26] and its cost $C_{new}$ can be calculated by function *Cost* (Line 3-6 of Algorithm 1). The function *Cost* is expressed as Eq. (20-22).

$$\boldsymbol{\delta} = [\delta_{dis}, \delta_{dir}, \delta_{mag}] \tag{20}$$

$$\boldsymbol{\mathcal{F}_{md}} = [Dis, \mathcal{F}_{dir}, \mathcal{F}_{mag}]^T \propto \boldsymbol{V_{gC}} \tag{21}$$

$$C = \boldsymbol{\delta}\, \boldsymbol{\mathcal{F}_{md}} \tag{22}$$

where $\boldsymbol{\delta}$ denotes the coefficient vector, $\propto$ denotes the dependency relation between $\mathcal{F}_{md}$ and $V_{gC}$. $Dis$ is the Euclidean distance between $\mathcal{s}_{goal}$ and current $\mathcal{s}$, it can be obtained through $Dis = \|\mathcal{s}_{goal} - \mathcal{s}\|^2$. $\mathcal{F}_{dir}$ and $\mathcal{F}_{mag}$ stands for the cost depended on direction and magnitude of global vector field respectively. $\mathcal{F}_{dir}$ is defined as Eq.(23).

$$\mathcal{F}_{dir}(\boldsymbol{\varsigma}, \boldsymbol{V_{gC}}) = \int_0^1 \left( w - l \frac{\dot{\varsigma}(\sigma)}{\|\dot{\varsigma}(\sigma)\|^2} \cdot \frac{\boldsymbol{V_{gC}}(\varsigma(\sigma))}{\|\boldsymbol{V_{gC}}(\varsigma(\sigma))\|^2} \right) \|\dot{\boldsymbol{\varsigma}}(\sigma)\|^2 d\sigma \tag{23}$$

where $\sigma$ is variable and $\sigma \in [0,1]$, $w$ and $l$ is constant coefficient. $\boldsymbol{\varsigma}$ denotes the path between two states. In VMD-RRT* algorithm without considering kinematic constraints, let $\boldsymbol{\varsigma}(0) = \mathcal{s}_{new}, \boldsymbol{\varsigma}(1) = \mathcal{s}_{nearest}$ *or* $\mathcal{s}_{near}$. $\mathcal{F}_{mag}$ is as Eq. (24).

$$\mathcal{F}_{mag} = \ln \boldsymbol{V_{gC}}(\boldsymbol{\varsigma}(\sigma)) \tag{24}$$

where $\sigma = 1$, $ln$ is logarithmic function with base $e$. When $C_{new}$ is calculated, the $\mathcal{s}_{new}$ and $r$ is set as center and radius of circle respectively. The state of $\mathcal{V}$, located within the range of this circle, is stored in $\mathcal{S}_{near}$ (Line 7-10 of Algorithm 1). Subsequently, let $\mathcal{s}_{min}$ and $C_{min}$ represent the state with the lowest cost and the corresponding cost value. It needs to find states stored in $\mathcal{S}_{near}$ to compare current $\mathcal{s}_{min}$ and $C_{min}$, then update the $\mathcal{s}_{min}$ with lower cost and its cost value (Line 12-18 of Algorithm 1). Both $\mathcal{s}_{min}$ and $\mathcal{s}_{new}$ are checked for their existence in area of obstacle and social gaze space. When $\mathcal{s}_{min}, \mathcal{s}_{new} \in \mathcal{S}_{free}$, $\mathcal{s}_{new}$ is considered as the new state in path $\boldsymbol{\mathcal{V}}$ and $\mathcal{s}_{min}, \mathcal{s}_{new}$ are added to $\boldsymbol{E}$, with $\mathcal{s}_{min}$ defined as the parent state of $\mathcal{s}_{new}$. Then, the algorithm will rewire for the tree $\boldsymbol{\Gamma}$ (Line 17-21 of Algorithm 1). When $\mathcal{s}_{new} \in \mathcal{S}_{goal}$, the path connecting $\mathcal{s}_{start}$ and $\mathcal{s}_{goal}$ (or $\mathcal{S}_{goal}$) can be found. The algorithm returns the expanded tree $\boldsymbol{\Gamma}$. The number of iterations reaches the threshold and failure is reported (Line 22-26 of Algorithm 1).

**Algorithm 1**: VMD-RRT*

**Input**: $\boldsymbol{u_{vmd}}$
**Output**: $\boldsymbol{\Gamma}$
1: $\boldsymbol{\mathcal{V}} \leftarrow \{\mathcal{s}_{start}\}$, $\mathbf{E} \leftarrow \emptyset$, $\boldsymbol{\Gamma} \leftarrow (\mathcal{V}, \mathrm{E})$;
2: **for** $\boldsymbol{i = 1 \dots n}$ **do**
3: $\boldsymbol{\mathcal{s}_{rand}} \leftarrow$ *RandomSampling*$(\boldsymbol{M_s})$;
4: $\boldsymbol{\mathcal{s}_{nearest}} \leftarrow$ *Nearest*$(\boldsymbol{\mathcal{V}}, \boldsymbol{\mathcal{s}_{rand}})$;
5: $\boldsymbol{\mathcal{s}_{new}} \leftarrow$ *Steer*$(\boldsymbol{\mathcal{s}_{nearest}}, \boldsymbol{\mathcal{s}_{rand}}, \boldsymbol{\varepsilon})$;
6: $\boldsymbol{C_{new}} \leftarrow$ *Cost*$(\boldsymbol{\mathcal{s}_{nearest}}, \boldsymbol{\mathcal{s}_{new}}, \boldsymbol{\mathcal{F}_{md}}, \boldsymbol{\mathcal{s}_{goal}}) + \boldsymbol{C_{nearest}}$;
7: **for** $\boldsymbol{j = 1 \dots length(\mathcal{V})}$ **do**
8: **if** $\|\boldsymbol{\mathcal{s}_{new}} - \boldsymbol{\mathcal{V}_j}\|^2 \le \boldsymbol{r}$
9: $\boldsymbol{\mathcal{S}_{near}} \leftarrow \boldsymbol{\mathcal{S}_{near}} \cup \{\boldsymbol{\mathcal{V}_j}\}$;
10: **end if**
11: **end for**
12: $\boldsymbol{\mathcal{s}_{min}} \leftarrow \boldsymbol{\mathcal{s}_{nearest}}$; $\boldsymbol{C_{min}} \leftarrow \boldsymbol{C_{new}}$;
13: **for** $\boldsymbol{k = 1 \dots length(\mathcal{S}_{near})}$ **do**
14: **if** $(\boldsymbol{C_{near_k}} +$ *Cost*$(\boldsymbol{\mathcal{S}_{near_k}}, \boldsymbol{\mathcal{s}_{new}}, \boldsymbol{\mathcal{F}_{md}}, \boldsymbol{\mathcal{s}_{goal}})) \le \boldsymbol{C_{min}}$
15: $\boldsymbol{\mathcal{s}_{min}} \leftarrow \boldsymbol{\mathcal{S}_{near_k}}$;
16: $\boldsymbol{C_{min}} \leftarrow \boldsymbol{C_{near_k}} +$ *Cost*$(\boldsymbol{\mathcal{S}_{near_k}}, \boldsymbol{\mathcal{s}_{new}}, \boldsymbol{\mathcal{F}_{md}}, \boldsymbol{\mathcal{s}_{goal}})$;
17: **end if**
18: **end for**
17: **if** *CollisionFree* $(\boldsymbol{\mathcal{s}_{new}}, \boldsymbol{\mathcal{s}_{min}})$
18: $\boldsymbol{\mathcal{V}} \leftarrow \boldsymbol{\mathcal{V}} \cup \{\boldsymbol{\mathcal{s}_{new}}\}$;
19: $\mathbf{E} \leftarrow \mathbf{E} \cup (\boldsymbol{\mathcal{s}_{new}}, \boldsymbol{\mathcal{s}_{min}})$;
20: $\boldsymbol{\Gamma}$.rewire();
21: **end if**
22: **if** $\boldsymbol{\mathcal{s}_{new}} \in \boldsymbol{\mathcal{S}_{goal}}$
23: **return** $\boldsymbol{\Gamma}$;
24: **end if**
25: **end for**
26: **return** failure;

3) **Auto-cruise of robot in social scenes**: The features of scene map are defined as $\Lambda_s = \{P, R, OBS, size\}$, $size$ is the size of map. The auto-cruise framework is showed in Fig. 6. It is composed of three components: Map feature extraction, Heuristic path and Motion planning. When the robot platform receives $\Lambda_s$, the OPP set $\mathbf{O}$ and superpositional social vector field $\boldsymbol{V_{gC}}$ will be derived from map feature extraction module (Part A, Part B in this section). The static obstacle area $\mathcal{S}_{obs}$ is delineated by the locations of obstacles. Besides, the group sequence $\chi$ is critical for robot auto-cruising in social scenes. In this paper, determining the cruising sequence among groups can be attributed to and combination optimization problem. This problem can be solved within TSP solver [36]. The effects from vector field of Magnetic Dipole $\boldsymbol{\mathcal{F}_{md}}$ can be calculated by $\boldsymbol{V_{gC}}$ and $\mathcal{S}_{free}$ that is generated through $\mathcal{S}_{free} = \mathcal{S} \backslash (\mathcal{S}_{obs} + \mathcal{S}_{SGS})$. Using the computed variables mentioned earlier, $\mathcal{S}_{free}$, $\boldsymbol{\mathcal{F}_{md}}$ and $\chi$ are utilized as the inputs of heuristic search module. Based on the sequence $\chi$, the robot can determine the next interaction group $\mathcal{G}_I$ (the informed sampling process is implemented in an ellipse zone to reduce the search costs). Then, the VMD-RRT* planner can generate a heuristic path $\boldsymbol{\mathcal{V}}$ between robot's position and the $OPP$ of $\mathcal{G}_I$, considering obstacles and the effects of the proposed vector field. By employing both the path $\boldsymbol{\mathcal{V}}$ and the robot motion space $\mathcal{S}_{rob}$ which is defined by a circle, where the radius of $\mathcal{S}_{rob}$ is expressed by $r_{rob}$, the sub-goals are determined by the intersection points of the path and robot motion space.

With the updating of sub-goals, sub-path $\boldsymbol{\mathcal{V}_s}$ is extracted from the heuristic path $\boldsymbol{\mathcal{V}}$. This sub-path $\boldsymbol{\mathcal{V}_s}$ is regard as the input of path tracking method. Through the path tracking calculation, the control commands will be generated to drive the robot approaching the sub-goal. As robot moving, the global positioning system continuously measures the states of scenario. Then, the robot updates $\Lambda_s$ and iterates the process to determine the next group to be approached and be interact.

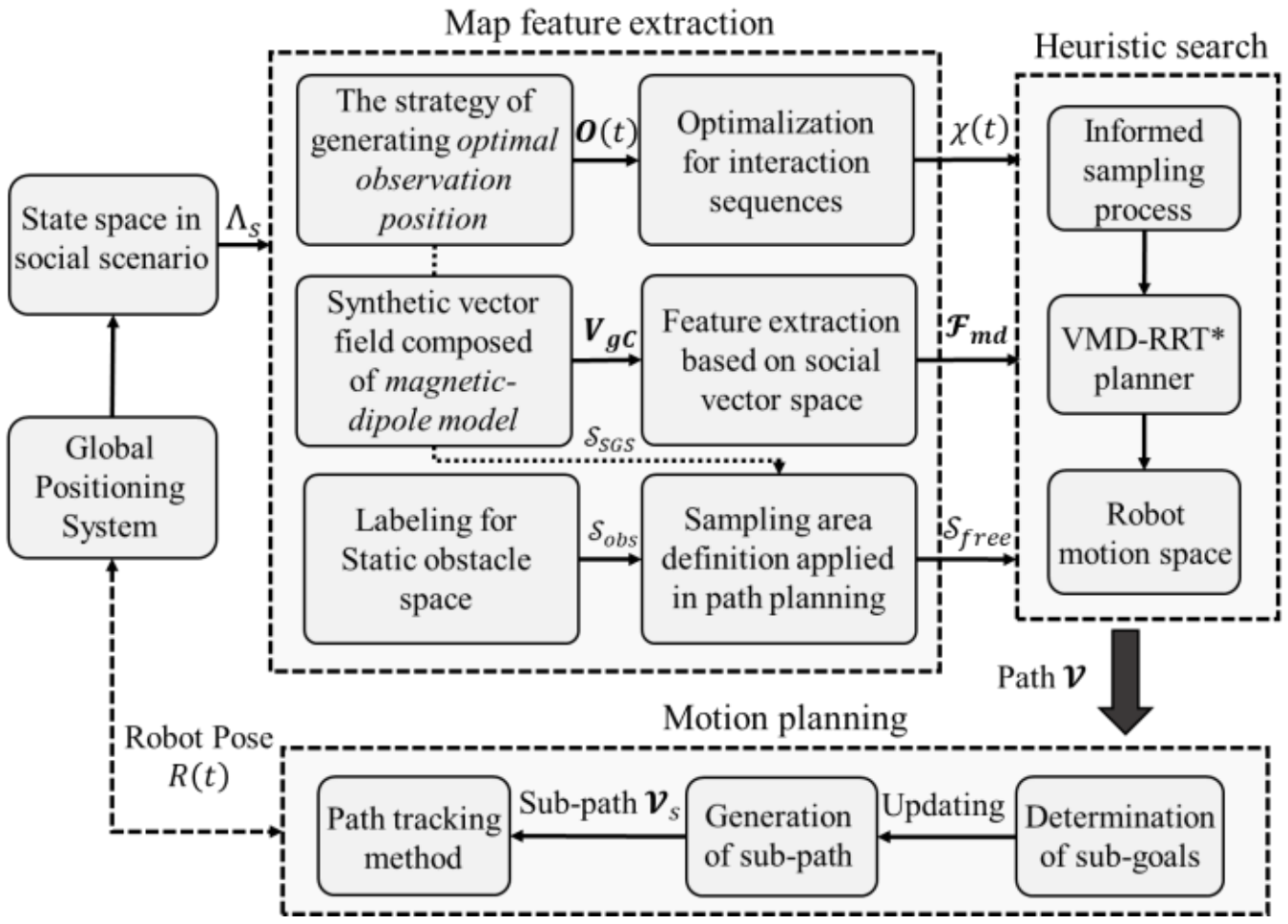

Fig. 6. The algorithm framework of robot Auto-cruise in social scenes.

## III. Simulation

In this section, simulation experiments are conducted to validate the effectiveness of proposed framework.

### *A. Simulation setup*

The simulation is conducted on the laptop with Intel(R) Core (TM) i5-1135G7 CPU. The map is created within Motion

Planning Tool Box of MATLAB, as shown in Fig. 7. The white space and grey space represent the free space and obstacle space, respectively (map range is 14 m × 14 m). In the simulated scenario, eight static individuals and two moving pedestrians are located. The blue circle and black circle represent the initial states of robot and individuals.

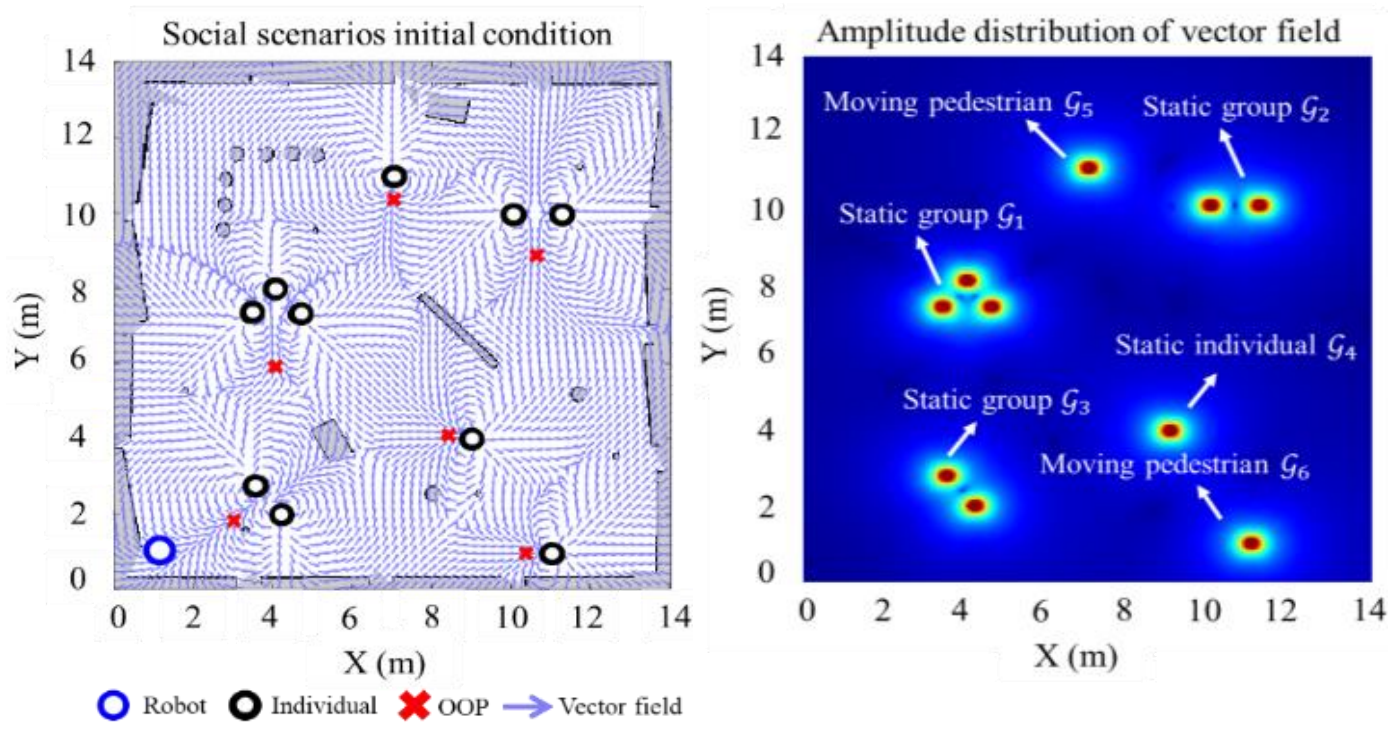

(a) (b)

Fig. 7. The initial state of designed social scenario, it contains both static and dynamic individuals. (a) the simulated map along with directional indicator of $\boldsymbol{V}_{gC}$. (b) the amplitude indicator of $\boldsymbol{V}_{gC}$

In Fig. 7(a), the blue arrow denotes the direction of current $\boldsymbol{V}_{gC}$ . Fig. 7(b) illustrates the amplitude of the $\boldsymbol{V}_{gC}$ , the increasing red color indicates higher amplitudes. Based on the clustering method proposed in Section II, the individual can be divided into six groups ($\mathcal{G}_1$~$\mathcal{G}_6$ in Fig. 7(b)). The purpose of this simulation is to make robot approach different groups based on the results of group clustering, and maintain a social compliant distance from each group during cruising. To initialize the simulation, some assumptions are made. *Assumption 1*: mobile robot can follow the heuristic path perfectly. *Assumption 2*: when robot approaching a person, the individual will notice the robot and wait for robot's arrival. *Assumption 3*: The kinematic constraints of robot are not considered. The parameters of robot motion planning in simulated scenario are listed in Table I.

TABLE I. PARAMETERS OF SIMULATION

| Items | Symbol | Value |
|---|---|---|
| The constant of social confidence | $a, b, c$ | 5.102,0.748,0.087 |
| The constant coefficient of Magnetic dipole-social space model | α | -615 |
| The coefficient of social damping | $\beta$ | -32 |
| The constant of gaze point | $\mathscr{r}$ | 0.05 |
| The scalar factor of Dis | $\delta_{dis}$ | 1 |
| The scalar factor of superpositional social vector field's direction cost | $\delta_{dir}$ | 5 |
| The scalar factor of superpositional social vector field's magnitude cost | $\delta_{mag}$ | 7 |
| The constant coefficient of $\mathcal{F}_{\text{dir}}$ | $w/l$ | 5/4 |
| Radius of robot motion space | $r_{rob}$ | 1.9 |
| Minimum distance to disturb the origin state of group | $D_{thres}$ | 0.4 |

*B. Simulation results and statistics analysis*

As shown in Fig. 8, the yellow star represents the OOP of specific group, and the green circles denote the sub-goals. Intuitively, the robot approaches different group OOPs according to the sequence $\chi$ determined by social proxemics field. The phenomenon of simulation can preliminarily demonstrate the effectiveness of proposed methodology. And the amplitude distribution of global proxemics field will be changed by moving pedestrians. The variation of the amplitude distribution affects the navigation process of robot. It can be found that robot tends to move in the direction perpendicular to the field direction in the motion planning procedure. On the other hand, robot would not cross into social gaze space to disrupt the origin interaction of group, this motion behavior basically satisfies the exception of robot planning tasks. Fig. 9 shows several critical indexes in robot motion planning. In Fig. 9(a), the blue line and red line represent the threshold to disturb the origin state of group and distance between robot and nearest individual respectively. According to [37], the threshold is set at 0.4m in compliance with the definition of social distance. In Fig.9 (a), it can be found that robot can maintain safe comfortable distance from different groups, and keep distance from nearest individual greater than $D_{thres}$. In Fig.9 (b), the moments for dynamic determination are recorded. $\mathcal{T}_{ddm}$ represents the trigger event of dynamic determination. When $\mathcal{T}_{ddm} = 1$, it means that the group that robot tend to approach has changed at that moment. And the decision behavior happened at every sub-goal. It indicates that the proposed algorithm can make real-time decision based on social proxemics feature. When robot is approaching a moving pedestrian, it needs to continuously adjust its motion as the pedestrian moving. To verify the stability and adaptability of proposed methodology, the simulation is conducted in four specified scenarios with the same map and different individual states. If the robot crosses into social gaze space or collides into individuals, it means the test of online motion planning fails. The simulation results are listed in Table II. Table II illustrates the planning performance evaluation when robot cruises in different social scenario. Where $n_{Sta}$ and $n_{Dyna}$ represent the number of static and dynamic individuals, respectively. The success rate indicates the probability that robot can successfully reach the OOPs of all the group. Intuitively, the robot can cruise and approach different groups with a high success rate in completely static scenario (the results are listed in the first two data columns in Table II). But as group pattern changes, the performance of proposed method becomes more unstable, which may result in planning failures (as listed in the last two data columns in Table II). It can be found that the path length is stable in the same scenario. The other parameters (average running time, and average exploration node) will increase as the increasing scene changing. In addition, with robot motion space expanding, it can be found that all indexes will decrease obviously. Generally, when $r_{rob} = 1.9m$, the stability and performance of proposed methodology can be maintained in the basically satisfactory level. Likewise, $r_{rob} = 1.9m$ will be applied in section IV.

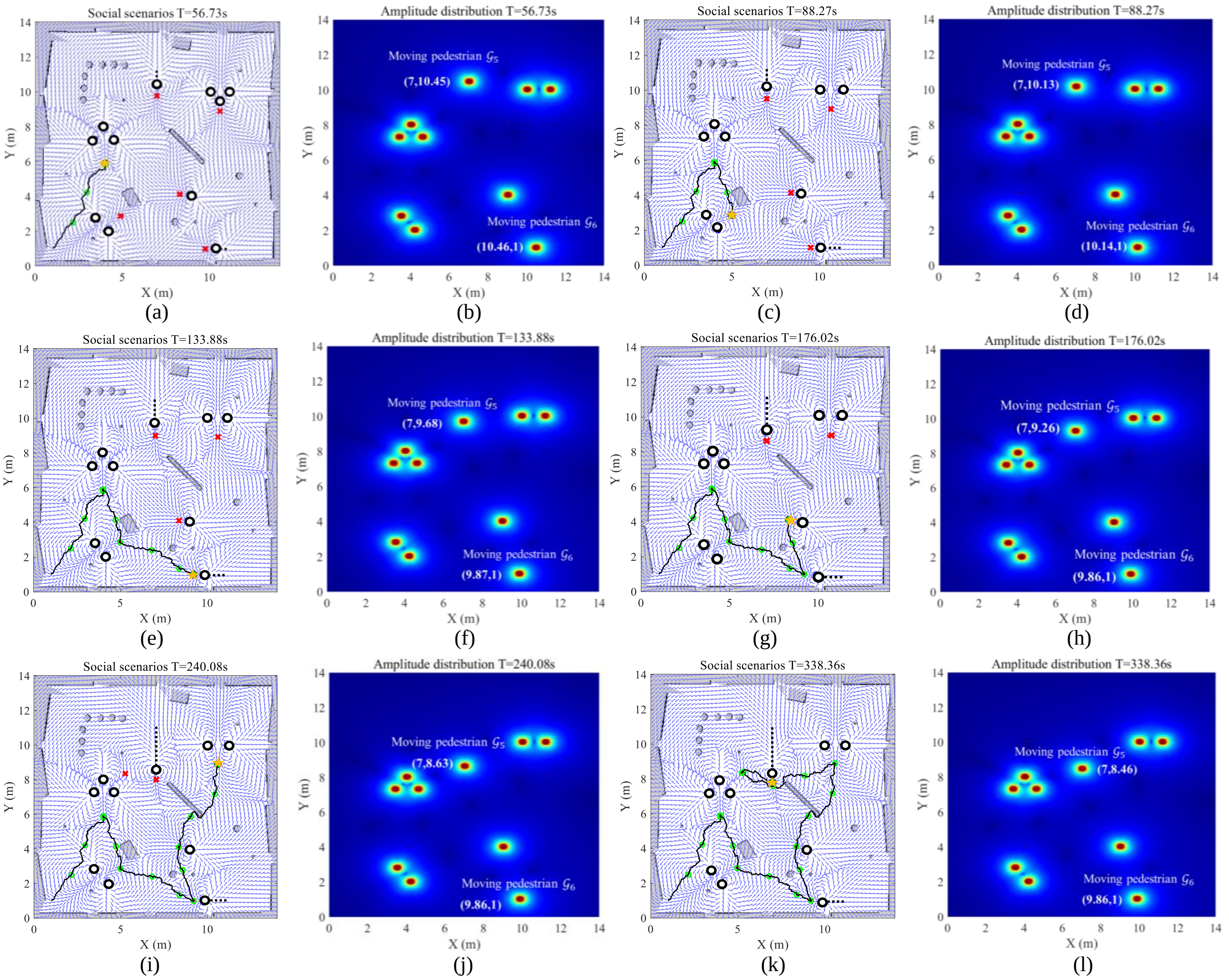

Fig. 8. Overall performance of proposed framework in designed social scenarios. The black line represents the sub-path generated by the robot Auto-Cruise Algorithm. The green circle shows the updating sub-goal. The dynamic individual's motion trajectory marked by the black dotted line. The yellow star represents the current goal for robot.

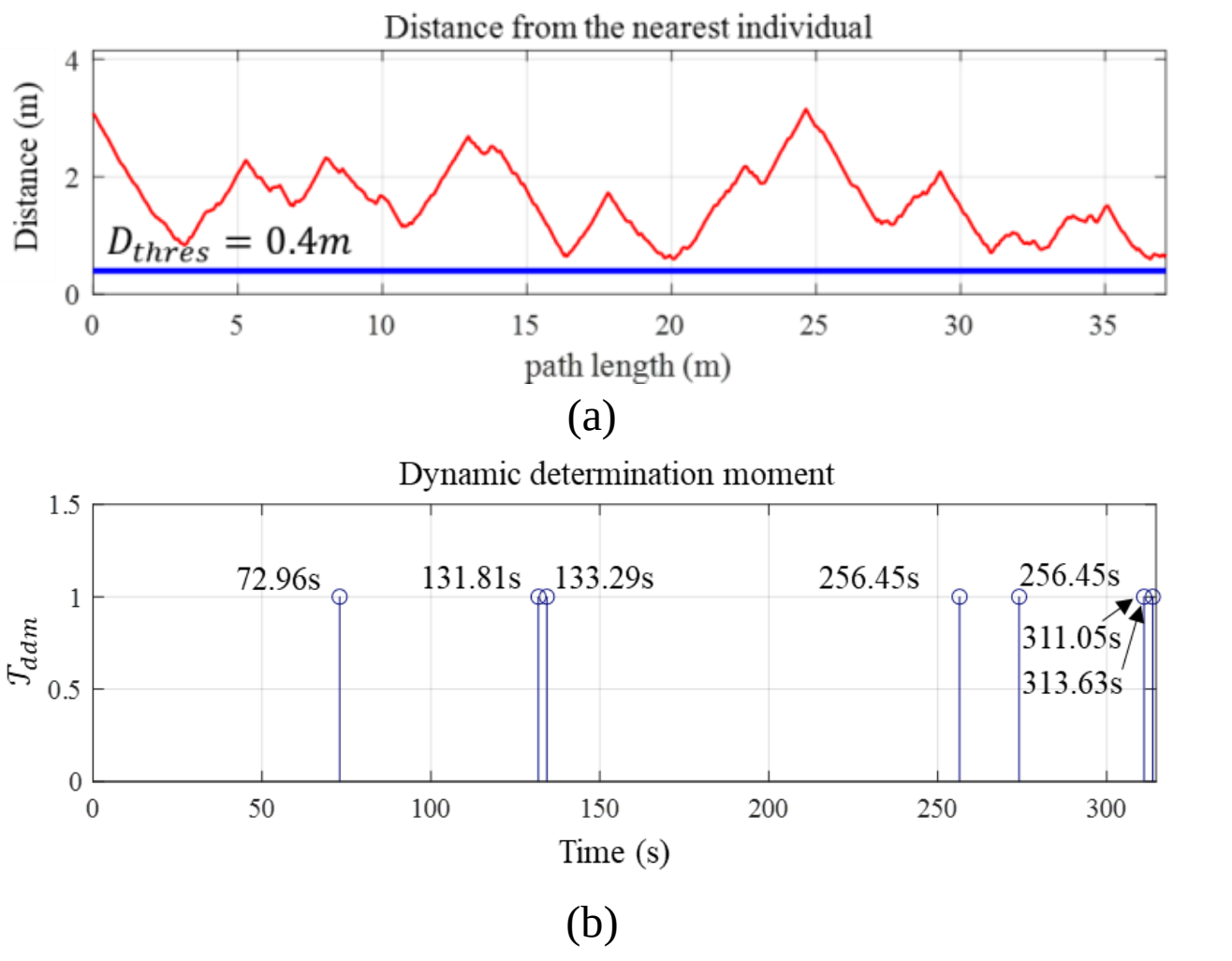

Fig. 9. (a) Distance between the robot and nearest individual. (b) Dynamic determination moment of interaction sequence.

## IV. Experiments and Result Analysis

### A. Experiment Setup

In the practical experimental studies, several volunteers are invited to be located in the test scene, as shown in Fig. 10. The system of setup is composed of three parts: computer (Intel NUC), motion capture system and service robot. The Xiaopang robot is employed as the moving platform. This robot has the complete movement device and software control interface connected to NUC computer. As for motion capture system, Vicon system is employed to capture the movement of volunteers and robot in real-time. Vicon is one kind of optical motion measurement system, it is equipped with camera array (eight high-speed cameras) to track the coordinates of markers (optical balls). During the experiments, volunteers wear a hat with six markers. Through tracing the position of theses markers, the three-dimensional states of these moving volunteers can be constructed. In this experimental setup, NUC computer Vicon system and robot constitute a local communication network (Ethernet protocol). NUC computer works as the control hub. It receives the volunteers and robot positions from the Vicon system, then conduct the calculation about volunteer states construction, proxemics field building and robot path-planning. The generated path, in the form of the control commands, will be sent to Xiaopang robot and guide it cruising in the experiment scene. The experiment scene is shown as Fig. 10, two figures in the lower left corner of Fig. 10 are the initial states of experimental social scenario. The parameters of experimental setup are identical to those in simulation section.

TABLE II
PERFORMANCE OF SIMULATION FOR DIFFERENT ROBOT MOTION SPACE AND SOCIAL SCENARIOS

| Robot motion space | | Social scenarios | | | |
|---|---|---|---|---|---|
| | | $n_{Sta}$: 6, $n_{Dyna}$: 0 | $n_{Sta}$: 10, $n_{Dyna}$: 0 | $n_{Sta}$: 6, $n_{Dyna}$: 1 | $n_{Sta}$: 6, $n_{Dyna}$: 2 |
| $r_{rob} = 1.5m$ | Success Rate | 10/10 | 10/10 | 10/10 | 10/10 |
| | Average Time(s) | 142.73 | 348.11 | 220.92 | 262.78 |
| | Path Length(m) | 26.58±4.96 | 44.07±8.00 | 34.59±7.48 | 38.26±7.23 |
| | Average Node | 7692 | 12638 | 10241 | 11356 |
| $r_{rob} = 1.9m$ | Success Rate | 10/10 | 10/10 | 10/10 | 10/10 |
| | Average Time(s) | 139.23 | 273.87 | 175.44 | 231.36 |
| | Path Length(m) | 26.89±4.38 | 42.16±6.95 | 36.37±6.82 | 40.33±6.19 |
| | Average Node | 6525 | 9573 | 8543 | 9108 |
| $r_{rob} = 2.3m$ | Success Rate | 10/10 | 10/10 | 9/10 | 8/10 |
| | Average Time(s) | 121.36 | 264.44 | 161.74 | 205.33 |
| | Path Length(m) | 25.14±6.23 | 43.41±6.22 | 33.94±6.58 | 38.61±6.81 |
| | Average Node | 5150 | 9573 | 7317 | 7637 |
| $r_{rob} = 2.7m$ | Success Rate | 10/10 | 10/10 | 7/10 | 4/10 |
| | Average Time(s) | 114.70 | 260.83 | 155.94 | 173.09 |
| | Path Length(m) | 24.98±6.33 | 44.94±8.85 | 35.91±7.27 | 40.20±5.12 |
| | Average Node | 4749 | 9139 | 6524 | 7059 |

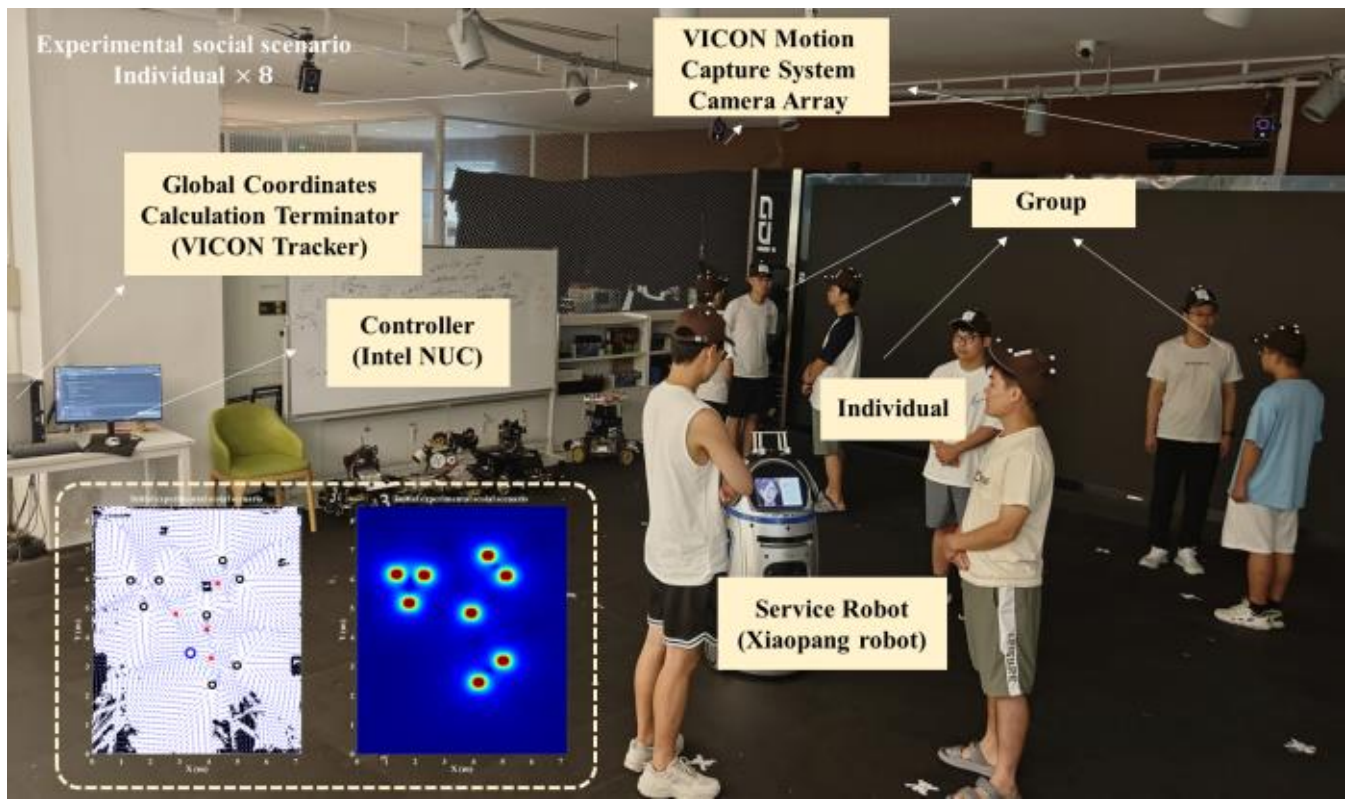

Fig. 10. The schematic diagram of experimental setup and actual scene.

*B. Static Scenario Experiments Analysis*

In the static scenario, we conducted robot auto-cruising among eight static volunteers. These volunteers form three multiple-person group and one individual group. Robot is designed to cruise in this scene. To evaluate whether the robot arrives at OOP and maintains a socially compliant orientation, Arrive Rate $A_r$ that considers both position and heading is introduced for evaluation, $A_r$ is defined as Eq. (25).

$$A_r = e^{-\left((x_r - O_x)^2 + (x_y - O_y)^2 + \sin^2(\theta_r - Arg(\xi_k) - \pi)\right)} \quad (25)$$

According to above definition, higher value of $A_r$ indicates that the robot has demonstrated the excepted pose in moving process. When the value of $A_r$ is higher than 90%, it is assumed that the robot's movement meet the satisfaction. The experiment results are demonstrated in Fig. 12. In Fig. 12(a), the black line with inverted green triangle denotes the robot trajectory in the experiments. It can be found that robot can reach all OOPs as the sequence $\chi$ obtained by TSP solver. Meanwhile, robot intends to maintain a safe distance from different groups in the cruise process. To evaluate the performance of motion planning module, the entire path is divided into four parts: N1, N2, N3, and N4. The robot trajectories in abovementioned four parts are shown in Fig. 11, where the blue solid line and dotted line denotes the $x$ coordinates and $y$ coordinates. The orange solid line denotes the heading of robot. In Fig. 11(a), the robot primarily takes about seven seconds to complete initialization. For N1~N4, each process has two phases, path planning phase and movement to the next OOP. In Fig. 11(b), when $O_2$ exceeds the range of robot motion space, the robot will arrive at current sub-goal and re-plan the path to $O_2$. These behaviors meet our assumption for robot auto-cruise in Section II. Additionally, several indicators, including path length, running time, and Arrive Rate $A_r$, are shown in Fig. 13. It can be found that the change of time and path length is within the allowed range (path length variation < 1m, running time variation < 15s), the $A_r$ generally greater than 90%. These statistic results demonstrate the stability of proposed methodology.

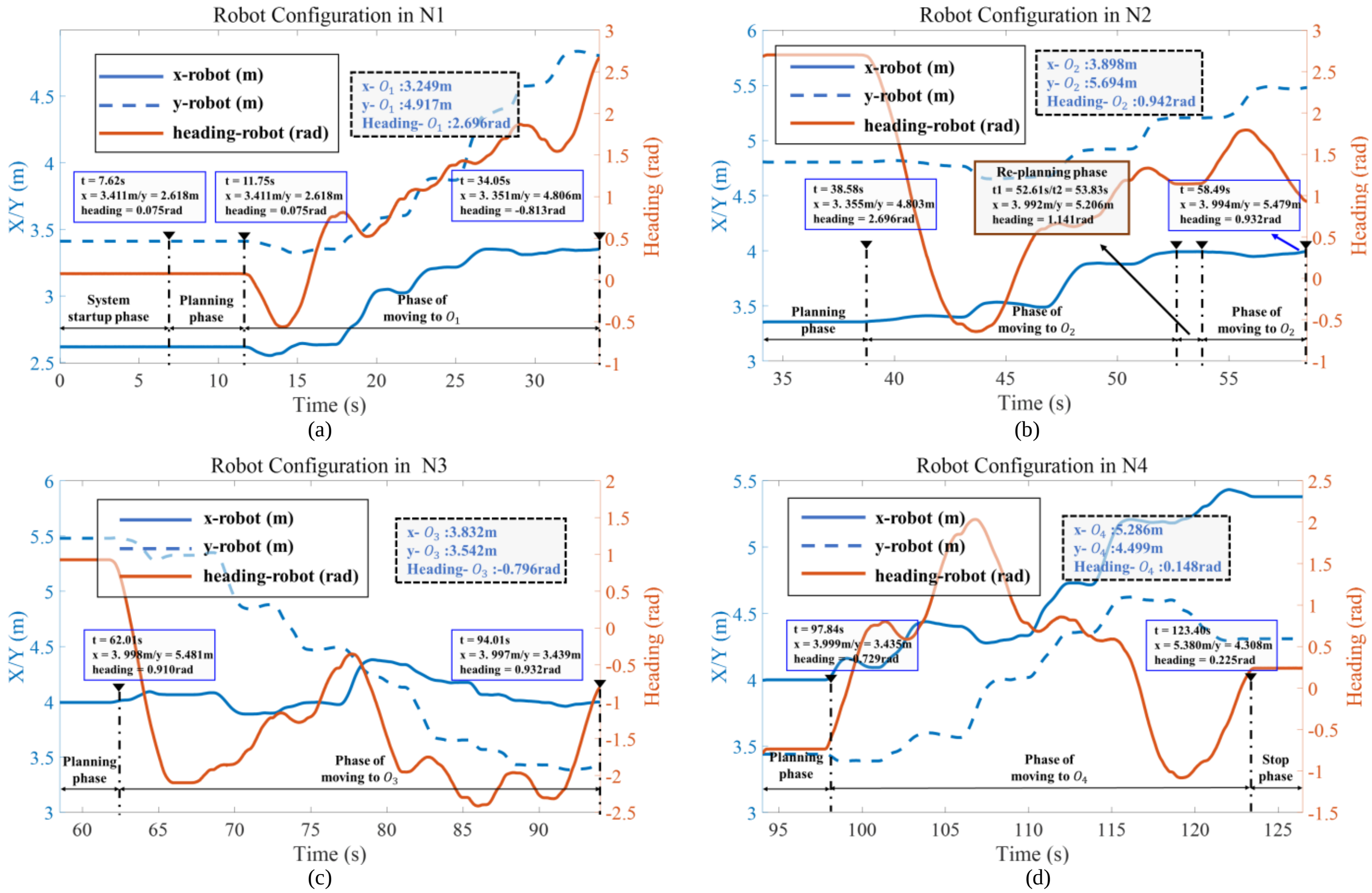

Fig. 11. The trace of robot movement in different part. The experiment is continuous process that contains system startup phase, planning phase, re-planning phase and phase of moving to next OOP.

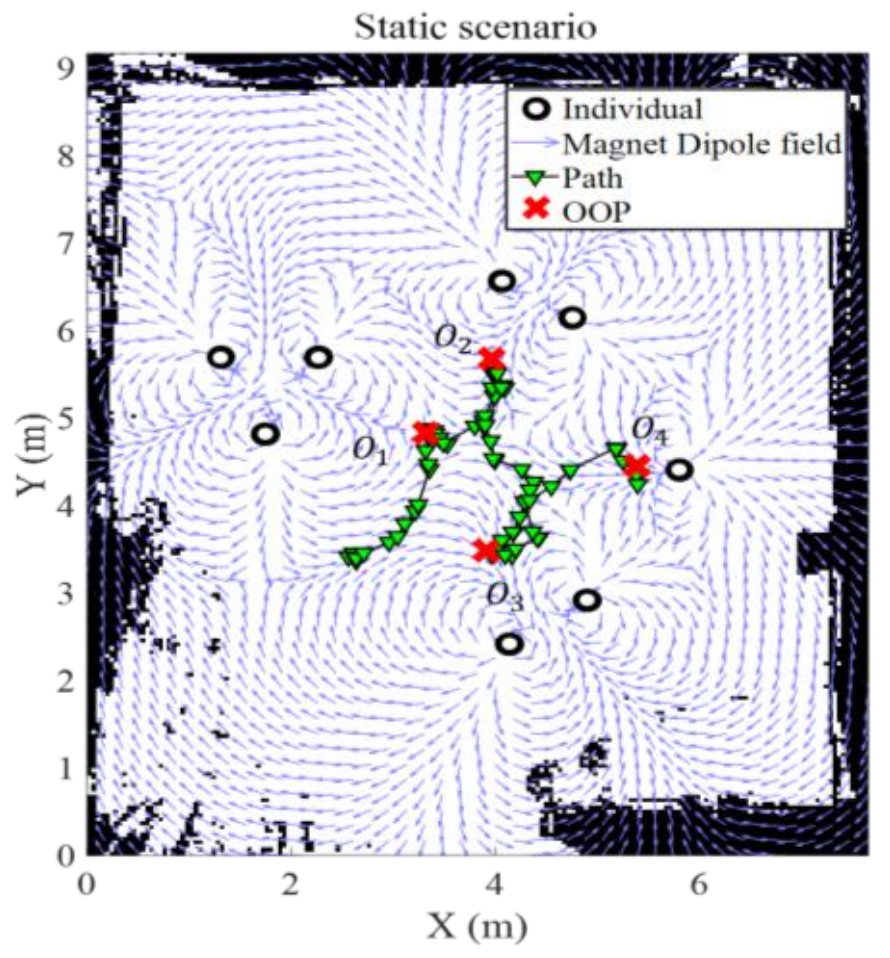

(a)

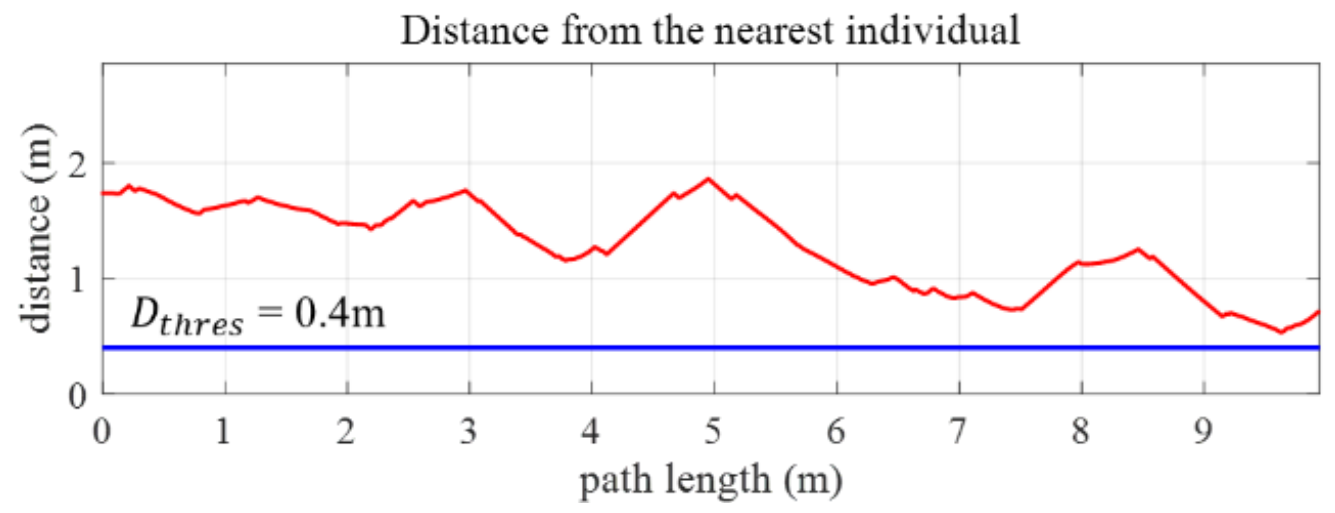

(b)

Fig. 12. (a) Overall performance in real-world experimental social scenarios. (b) Distance between the robot and nearest individual.

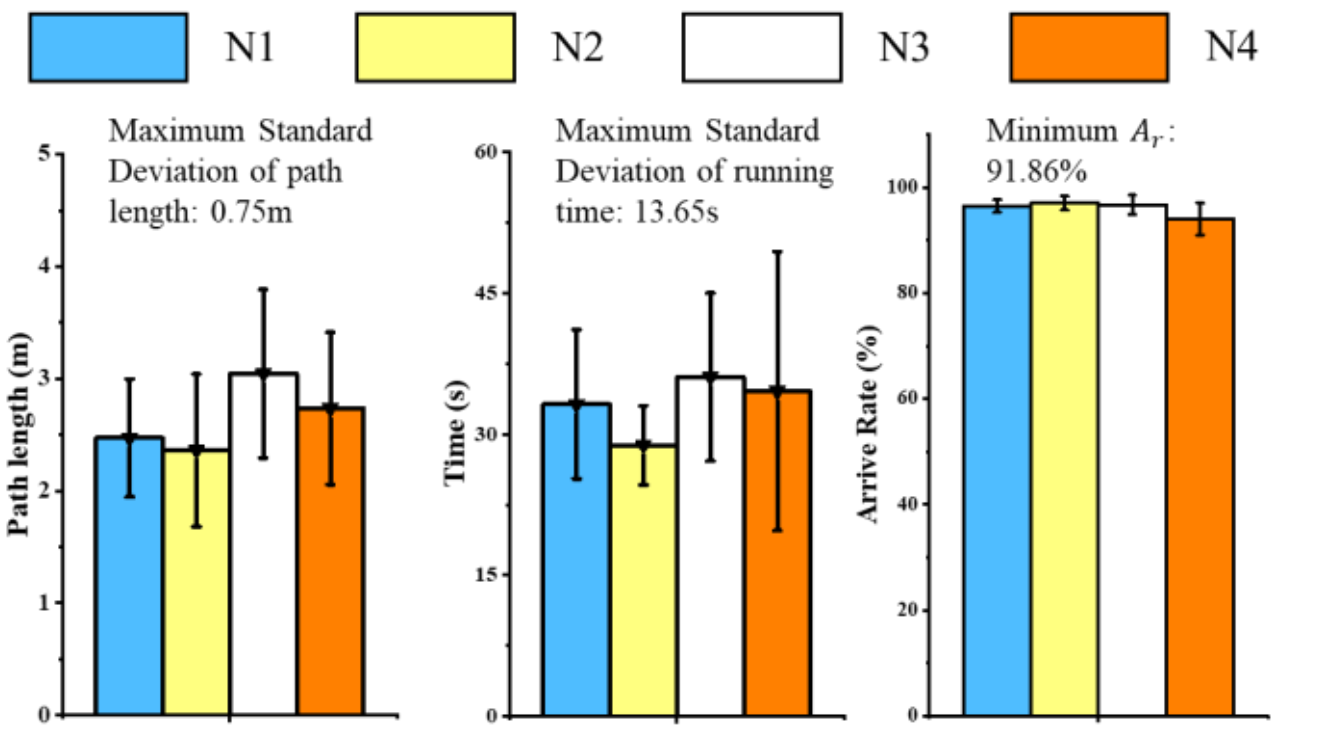

Fig. 13. Statistics analysis of critical index for each motion planning part in experimental social scenario.

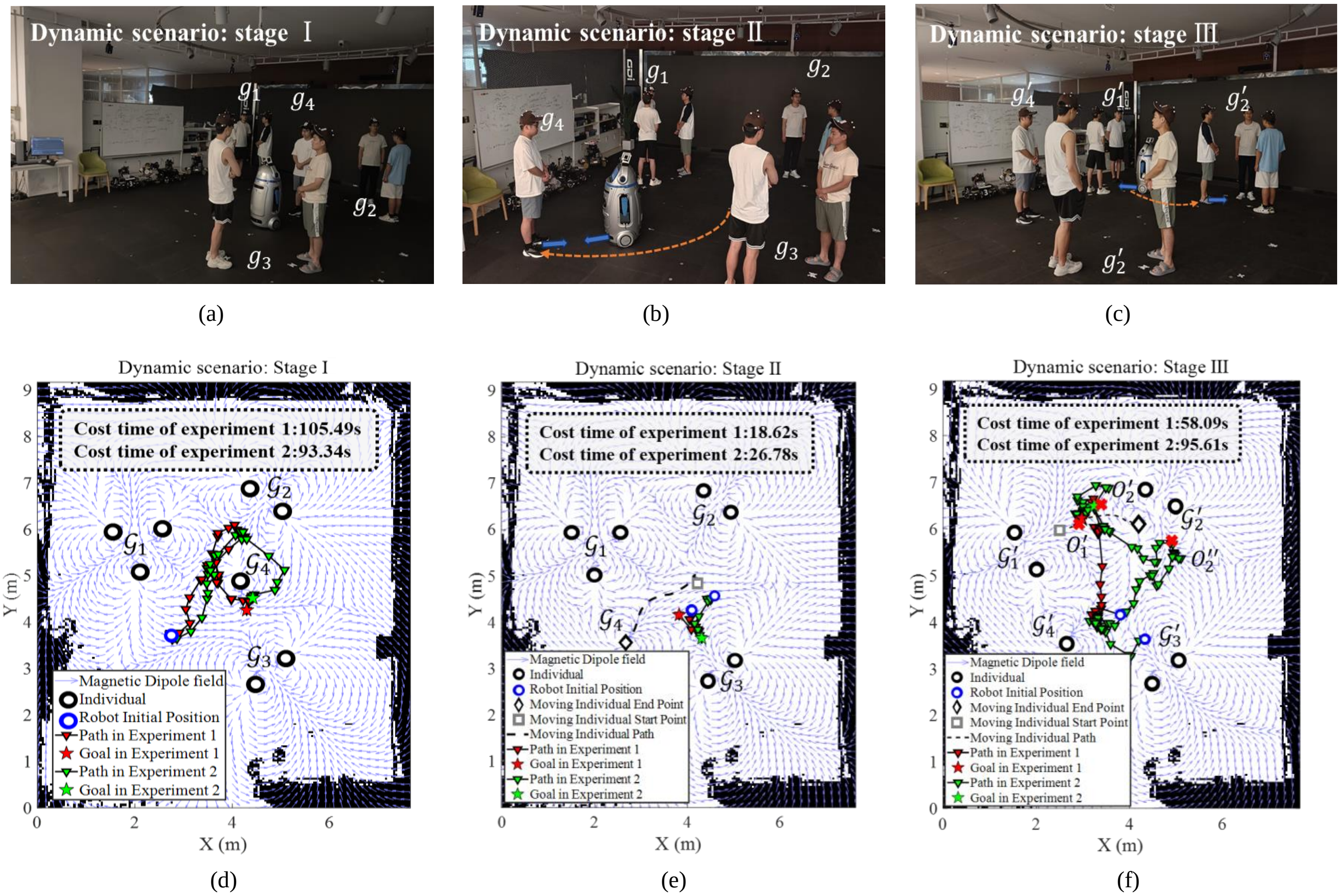

Fig. 14. Display of robot auto-cruise in dynamic social scenario applies our proposed framework. (a)(b)(c) show the real scene in different stage of current experiment. (d)(e)(f) demonstrate the robot's motion planning in response to changes in group dynamics.

*C. Dynamic Scenario Analysis*

To verify the adaptability of proposed methodology in dynamic social scenario, we have organized the experiments shown as Fig. 14. In this dynamic experiment, there will exist the change of the grouping pattern. Initially, four groups ($\mathcal{G}_1 \sim \mathcal{G}_4$ in Fig. 14(a)) are standing in the test scene. When robot is cruising among them, volunteers will change their positions and grouping pattern at a random moment. According to the occurrence of the grouping reorganization, the dynamic experiment is divided into three phases (Fig. 14). In the stage I, the robot extracts proxemic features from the current scene and is ready to approach groups. In the stage II, individual group $\mathcal{G}_4$ moves to specific position. The final states of stage II is shown in Fig. 14(b) where the orange dotted line represents the approximate moving trajectory of $\mathcal{G}_4$ and the bule arrow represents the heading of $\mathcal{G}_4$. The service robot has to perceive the change of scene and adjust its interactive cruising trajectories. In the stage III, an individual from $\mathcal{G}_1$ starts to move towards $\mathcal{G}_2$ and interact with $\mathcal{G}_2$. Finally, the new groups ($\mathcal{G}'_1 \sim \mathcal{G}'_4$ in Fig. 14(c)) are generated by developed clustering method. When the groups reconstitute, the social group proxemic feature of current scene alters, and the robot adjusts the changes to cruise among the new groups again.

To validate the repeatability of proposed methodology in real-world scenario, two experiment are conducted (experiment 1 and experiment 2 in the second row of Fig. 14). In Fig. 14(d), the robot starts to cruise and approach $\mathcal{G}_1$, $\mathcal{G}_2$, $\mathcal{G}_4$ according to the initial scene setup in two experiments. Then, as $\mathcal{G}_4$ moves as shown in Fig. 14(e), the robot tends to reapproach $\mathcal{G}_4$ after interacting with $\mathcal{G}_3$. In Fig. 14(f), the robot can adapt to the changes of groups and successfully approach to the new formed groups at generated OOPs. There are some phenomena can be observed between the two experiments. When grouping pattern changes, the two generated paths don't overlap, which may be caused by the quasi-optimality and randomness of VMD-RRT* algorithm. And due to the robot starts from initial positions in the two processes, the proposed framework will also generate various paths. Besides, the interaction sequence $\chi$ calculated by TSP solver differs, especially in Fig. 14(f). The reason for this phenomenon maybe is the distinct initial states of service robot and the non-optimality solution computed by TSP solver. In stage III, the interaction sequences for experiment 1 and experiment 2 were $\{O'_4, O'_2, O'_1\}$ and $\{O'_4, O''_2, O'_1, O'_2\}$, respectively. The service robot makes duplicate observation for $\mathcal{G}'_2$. It demonstrates that the generation of OOP would be various due to the robot configurations and provides the possibility for multi-faceted observation of group.

In order to evaluate whether the above characteristics of proposed method has an impact on the overall framework, we compare it with other global path planning algorithm and record the execution time, path length and traversed nodes in the same experimental scenario. The results are listed in TABLE III. The experimental data shows that the proposed framework achieves better performance and stability, with stable planning time, shorter path lengths and fewer nodes. Generally, these experiments can prove the effectiveness of developed method in this article.

TABLE III
COMPARISON OF EXPERIMENTAL DATA WITH OTHER METHOD

| Algorithm | Time (s) | Path Length (m) | Node |
|---|---|---|---|
| RRT | 187.76±25.57 | 33.75±5.30 | 15855±3754 |
| RRT* | 255.60±29.75 | 26.28±1.82 | 20063±1318 |
| **VMD-RRT*** | **192.03±16.03** | **23.93±1.06** | **6548±980** |

## V. CONCLUSION

In this paper, efforts are taken to blend the group interaction awareness into the robot navigation. In order to make robot identify the groups in dynamic and crowded scenario, we develop a novel group clustering method, and propose defining the proxemics of group within magnetic dipole model. On the basis of group clustering and proxemics modeling, a method to calculate the optimal observation positions (OOPs) of group is established. Through above works, the proxemics of people can be transformed into the global map composed of unique vector filed and OOPs grid, its mixture feature can naturally be utilized to guide robot cruising. Therefore, we propose a hierarchical path planning algorithm for the robot in the global map (containing static obstacle). Both simulation and physical experiments have been conducted based on one service robot platform, which can promisingly facilitate the development of robot public service. The further work is to improve the method performance in trapped and narrow environments, and combine the cruising behavior with the practical interaction behavior.

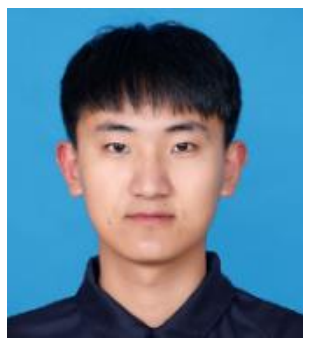

**Xuan Mu** received the B.S. in automation from the Qingdao University, Qingdao, China, in 2025. He is currently pursuing the master’s degree in Qingdao University, Qingdao, China. His research interests include robot control, combination optimization theory and intelligent human-robot interaction.

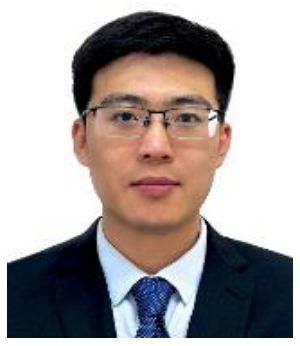

**Xiaorui Liu** received the M.Sc. and Ph.D. degree in intelligent information and communication system from Ocean University of China, in 2014 and 2019 respectively. In 2018, he participated the Electromagnetic Compatibility Lab of Missouri S&T as a visiting scholar. He is currently an associate professor with the school of automation, Qingdao University, China. His main research interests include control and instrumentation in the robotics, human-machine interaction, intelligent measurement system and deep learning. Dr. Liu is a member of Chinese Association of Artificial Intelligence and Chinese Association of Automation. In 2022, he received best paper award of IEEE International Conference on Advanced Robotics and Mechatronics.

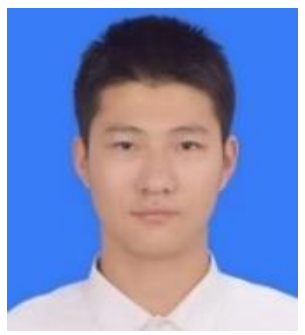

**Shuai Guo** received the B.S. in automation from the Qingdao University, Qingdao, China, in 2025. He is currently pursuing the master’s degree in Qingdao University, Qingdao, China. His research interests include unmanned aerial vehicle control, convex optimization theory and deep learning.

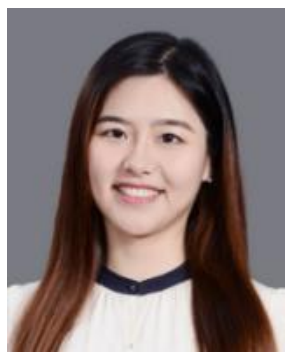

**Wenzheng Chi** received the B.Sc. degree in automation from Shandong University, Jinan, China, in 2013 and the Ph.D. degree in biomedical engineering from the department of electronic engineering, Chinese University of Hong Kong, Hong Kong, in 2017. During her Ph.D. degree, she spent six months at the University of Tokyo, Japan, as a visiting scholar. She was a postdoctoral fellow with the department of electronic engineering, Chinese University of Hong Kong, from 2017 to 2018. She is currently a professor with the robotics and microsystems center, school of mechanical and electric engineering, Soochow University, Suzhou, China. Her research interests include mobile robot path planning, intelligent perception, human-robot interaction, etc.

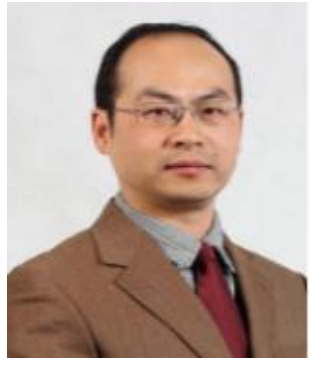

**Wei Wang** the B.Sc. and M.Sc. degree in mechanical design and manufacture from Harbin Engineering University in 1994 and 1997 respectively. Receive the Ph.D. degree in mechanical and electronic engineering from Beihang University in 2000. In 2003, he participated in Universität Hamburg as a visiting scholar. He is currently a professor of the school of mechanical engineering and automation, Beihang University. His research interests include mobile&service robot, human-robot interaction, robot emotion model etc.

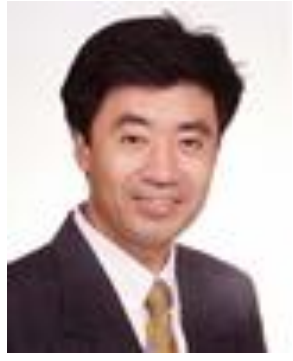

**Shuzhi Sam Ge** received the B.Sc. degree in control engineering from the Beijing University of Aeronautics and Astronautics, Beijing, China, in 1986 and the Ph.D. degree in mechani-cal/electrical engineering from the Imperial College London, London, U.K., in 1993. He is the Director with the Social Robotics Lab-oratory of Interactive Digital Media Institute, Singapore, and a Professor with the Department of Electrical and Computer Engineering, National University of Singapore, on leave from the director of the institute of future, Qingdao University, Qingdao. He has coauthored four books and over 300 international journal and conference papers. His current research interests include social robotics, adaptive control, intelligent systems, and artificial intelligence. Dr. Ge has served/been serving as an Associate Editor for a number of flagship journals, including IEEE TRANSACTIONS ON AUTOMATION CONTROL, IEEE TRANSACTIONS ON CONTROL SYSTEMS TECHNOLOGY, IEEE TRANSACTIONS ON NEURAL NETWORKS and Automatica. He served as the Vice President for Technical Activities, from 2009 to 2010, the Vice President of Membership Activities, from 2011 to 2012, and a member of the Board of Governors, from 2007 to 2009 at the IEEE Control Systems Society. He is a Fellow of the International Federation of Automatic Control, the Institution of Engineering and Technology, and the Society of Automotive Engineering.